\documentclass{article}

\usepackage{microtype}
\usepackage{graphicx}
\usepackage{subfigure}
\usepackage{booktabs} 
\usepackage{tikz}
\usetikzlibrary{shapes,arrows,fit}

\usepackage[pagebackref]{hyperref}

\usepackage[accepted]{icml2024}

\usepackage{amsmath}
\usepackage{amssymb}
\usepackage{mathtools}
\usepackage{amsthm}

\usepackage[capitalize,noabbrev]{cleveref}

\theoremstyle{plain}
\newtheorem{theorem}{Theorem}[section]

\theoremstyle{definition}

\newtheorem{assumption}[theorem]{Assumption}
\theoremstyle{remark}
\newtheorem{remark}[theorem]{Remark}

\usepackage[textsize=tiny]{todonotes}

\icmltitlerunning{Stochastic Unrolled Federated Learning}

\begin{document}

\twocolumn[
\icmltitle{Stochastic Unrolled Federated Learning}



\icmlsetsymbol{equal}{*}

\begin{icmlauthorlist}
\icmlauthor{Samar Hadou}{Penn}
\icmlauthor{Navid NaderiAlizadeh}{Duke}
\icmlauthor{Alejandro Ribeiro}{Penn}
\end{icmlauthorlist}

\icmlaffiliation{Penn}{Department of Electrical and Systems Engineering, University of Pennsylvania, PA, USA}
\icmlaffiliation{Duke}{Department of Biostatistics and Bioinformatics, Duke University, NC, USA}

\icmlcorrespondingauthor{Samar Hadou}{selaraby@seas.upenn.edu}

\icmlkeywords{Machine Learning, ICML}

\vskip 0.3in
]



\printAffiliationsAndNotice{}  



\begin{abstract}
    Algorithm unrolling has emerged as a learning-based optimization paradigm that unfolds truncated iterative algorithms in trainable neural-network optimizers. We introduce Stochastic UnRolled Federated learning (SURF), a method that expands algorithm unrolling to federated learning in order to expedite its convergence. Our proposed method tackles two challenges of this expansion, namely the need to feed whole datasets to the unrolled optimizers to find a descent direction and the decentralized nature of federated learning. We circumvent the former challenge by feeding stochastic mini-batches to each unrolled layer and imposing descent constraints to guarantee its convergence. We address the latter challenge by unfolding the distributed gradient descent (DGD) algorithm in a graph neural network (GNN)-based unrolled architecture, which preserves the decentralized nature of training in federated learning. We theoretically prove that our proposed unrolled optimizer converges to a near-optimal region infinitely often. Through extensive numerical experiments, we also demonstrate the effectiveness of the proposed framework in collaborative training of image classifiers.
\end{abstract}

\section{Introduction}

Federated learning (FL) is a distributed learning paradigm in which a set of low-end devices aim to collaboratively train a global statistical model. 
A growing body of work, e.g., \citep{lian2015asynchronous, McMahan2016, Li2020On}, has deployed a server in the network to facilitate reaching consensus among the agents, which  creates a communication bottleneck at the server and requires high bandwidth when the number of agents grows large. 
To alleviate these challenges, another line of work that traces back to decentralized optimization \citep{nedic_distributed_2009, wei2012distributed, wu2017decentralized} has instead investigated peer-to-peer communication, eliminating the role of central servers in the network. These decentralized federated learning frameworks compromise communication efficiency and convergence rates \citep{vanhaesebrouck2017decentralized, Liu22com, liu2022decentralized}. The slow convergence of these methods arises as a practical challenge since it greatly outweighs the capacity of resource- and energy-constrained devices.

Algorithm unrolling has recently emerged as a learning-to-optimize paradigm that unfolds iterative algorithms via learnable neural networks, thereby enabling learning the parameters of the iterative algorithm. The layers of the unrolled architecture, also referred to as the unrolled optimizer, correspond to the iterations of the standard one while the outputs of these layers form a trajectory toward the optimal.
The key reported advantage of learning the parameters is achieving \emph{much faster convergence} \citep{ monga_algorithm_2021} while achieving state-of-the-art performance in many applications such as sparse coding \citep{gregor10}, computer vision \citep{zhang2018ista}, policy learning \citep{marino2021iterative}, and computational biology \citep{cao2019learning} to name a few. 

Nevertheless, a question arises concerning convergence: \emph{Are unrolled network guaranteed to converge?} 
Some studies, e.g., \citep{xie2019differentiable, chen2018theoretical}, provided theoretical proofs for the existence of unrolled networks that converge; however, they do not provide methods for finding these convergent networks.  To resolve this issue, \citep{liu2019alista, Abadi15} reduce the size of the search space by learning fewer parameters of the standard algorithm, which limits the network's expressivity. 
Another approach, known as safeguarding, has been proposed in \citep{Heaton_Chen_Wang_Yin_2023, shen2021learning, Moeller_2019_ICCV, liu2021investigating}, where the estimate made by a certain layer is considered only if it is in a descent direction; otherwise, it is replaced with an estimate of the classic iterative algorithm to guarantee convergence. More recently, \citep{hadou2023robust} has provided convergence guarantees, agnostic to the standard algorithms being unrolled, by forcing convergence constraints during training.

In this paper, we introduce unrolling to federated learning and provide convergence guarantees within a proposed training framework called Stochastic UnRolled Federated learning (SURF). Unlike previous works in unrolling that train unrolled architectures to solve an optimization problem, SURF targets empirical risk minimization (ERM) problems. This positions SURF as a train-to-train, or learn-to-learn (L2L), method that uses \emph{meta-training} to train the unrolled network. Consequently, an additional challenge emerges as we feed each entry in the meta-training dataset, which is a downstream dataset itself, into the unrolled architecture that has a fixed-size relatively-small input. To bypass this challenge, we introduce \emph{stochastic unrolling} within our framework, where we inject a mini-batch randomly sampled from the downstream dataset to each layer of the unrolled architecture. Stochastic unrolling, however, introduces uncertainty in the estimated descent directions during inference, therby adding another level of subtlety to its (or its lack of) convergence guarantees.
Similar to \citep{hadou2023robust}, SURF imposes \emph{descending constraints} on the training procedure of the unrolled architecture to guarantee convergence despite the introduced uncertainty. 

The second facet of our proposal involves crafting an unrolled architecture suitable for both decentralized and classical federated learning problems. To achieve this, we consider a general formulation that frames the FL problem as a distributed optimization problem. Hence, we construct our unrolled architecture by unrolling decentralized gradient descent (DGD) \citep{nedic_distributed_2009}, a well-established algorithm for distributed problems, with the help of graph neural networks (GNNs). We show that unrolled DGD (U-DGD), albeit designed for decentralized FL, can be extended to classical FL scenarios where a server node is deployed.

Lastly, we supplement our proposal with a theoretical analysis, demonstrating that unrolled architectures trained through SURF are stochastic descent optimizers, which converge to a near-optimal region of the FL loss function. Additionally, we prove that the unrolled network has \emph{exponential} convergence rates, surpassing the sublinear rates reported for the state-of-the-art FL methods.

In summary, our contributions are as follows:
\begin{itemize}
    \item We introduce SURF, an L2L method for federated learning empowered by algorithm unrolling.
    \item We unroll DGD in GNN-based unrolled layers that can handle both decentralized and classical FL problems.
    \item We allow whole datasets to be fed to U-DGD through stochastic unrolling and force the unrolled architecture to converge by imposing descending constraints.  
    \item We theoretically prove that an unrolled network, trained via SURF, converges to a near-optimal region with an exponential convergence rate.
\end{itemize}

One of the advantages that SURF provides is shifting where a neural network is trained using gradient descent, i.e., moving from training a neural network online to training an unrolled network offline. This moves the demanding hardware training requirements from low-end devices to more powerful offline servers. The downside of using unrolling in training neural networks is that the size of the unrolled network is typically much larger than the original one. Therefore, we envision SURF as a method that complements other federated learning frameworks without necessarily replacing them. Particularly, SURF best suits problems of training relatively lightweight models on resource- and energy-limited devices, 
where fast convergence is a priority.
\section{Related Work}

Algorithm unrolling aims to unroll the hyperparameters of a standard iterative algorithm in a neural network to learn them. The seminal work \citep{gregor10} unrolled iterative shrinkage thresholding algorithm (ISTA) for sparse coding problems. Following \citep{gregor10}, many other algorithms have been unrolled, including, but not limited to, projected gradient descent \citep{Giryes18}, the primal-dual hybrid gradient algorithm \citep{Jiu2020ADP}, and Frank-Wolfe \citep{liu2019frank}.

Algorithm unrolling has also been introduced to distributed optimization problems with the help of graph neural networks (GNNs). One of the first distributed algorithms to be unrolled was weighted minimum mean-square error
(WMMSE)~\citep{shi2011iteratively}, which benefited many applications including wireless resource allocation \citep{Chowdhury21, Li22} and multi-user multiple-input multiple-output (MU-MIMO) communications \citep{Hu21, zhou22, Ma22, Pellaco22, Schynol22, Schynol23}. Several other distributed unrolled networks have been developed for graph topology inference \citep{pu2021learning}, graph signal denoising \citep{Chen21denoising, Nagahama21} and computer vision \citep{lin2022ru}, among many others. In our work, we follow the lead of these studies and rely on GNNs to unroll DGD for federated learning. To the best of our knowledge, our work is the first to use algorithm unrolling in a federated learning setting. An extended version of Related Work can be found in Appendix \ref{app:related}.

\section{Background} \label{sec:PF}
We start our discussions with a brief review of algorithm unrolling and introduce a general FL formulation.
\subsection{Algorithm Unrolling} \label{sec:unrolling_basics}
The primary goal of unrolling is to learn to solve the optimization problem $\text{argmin}_{\bf z} f({\bf z}; {\bf u})$, usually solved by an iterative algorithm $\cal A$. The latter, randomly initialized by ${\bf z}_0$, executes an update rule of the form ${\bf z}_l = \phi({\bf z}_{l-1}, f; \boldsymbol{\theta})$ sequentially till it converges to a stationary point of the objective function $f$. The parameterization $\boldsymbol{\theta}$ of the update function $\phi$ is commonly crafted based on domain knowledge. In the simplest case of gradient descent, the hyperparameter is the step size, and the update rule is ${\bf z}_l = {\bf z}_{l-1} - \theta \nabla f({\bf z}_{l-1}; {\bf u})$.

In the unrolling process, the hyperparameters $\boldsymbol{\theta}$ of the standard algorithm $\cal A$ are set free to learn. The unrolled architecture unfolds $\cal A$ in $L$ neural layers, each of which resembles the update map $\phi$ with learnable parameters $\boldsymbol{\theta}_l$ that vary per layer. The output of each unrolled layer is fed to the next one along with the parameterization $\bf u$ since $f$, as a continuous operator, cannot be directly fed into the neural network. An illustration of the unrolled architecture is depicted in \cref{fig:unrolled_net}.
\tikzstyle{block} = [draw, fill=lightgray, rectangle, 
    minimum height=1em, minimum width=6em]
\tikzstyle{dots} = [rectangle, 
    minimum height=1em, minimum width=1em]
\tikzstyle{sum} = [draw, fill=white, circle, node distance=1cm]
\tikzstyle{input} = [coordinate]
\tikzstyle{output} = [coordinate]
\tikzstyle{pinstyle} = [pin edge={to-,thin,black}]
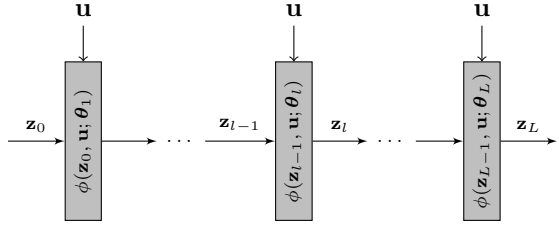
\begin{figure}[t]
    \centering
    \begin{tikzpicture}[auto, node distance=2cm,>=latex']

        \node [input, name=input] {};
        \node [block, right of=input, rotate=90, pin={[pinstyle]right:$\bf u$},
                node distance=1cm] (first) {\scriptsize$ \phi({\bf z}_{0}, {\bf u}; \boldsymbol{\theta}_1)$};
        \node [dots, right of=first,
                node distance=1.3cm] (dots1) {\scriptsize $\dots$};
        \node [block, right of=dots1, rotate=90, pin={[pinstyle]right:$\bf u$},
                node distance=1.5cm] (middle) {\scriptsize $\phi({\bf z}_{l-1}, {\bf u}; \boldsymbol{\theta}_l)$};
        \node [dots, right of=middle,
                node distance=1.3cm] (dots2) {\scriptsize $\dots$};
        \node [block, right of=dots2, rotate=90, pin={[pinstyle]right:$\bf u$},
                node distance=1.2cm] (last) {\scriptsize $\phi({\bf z}_{L-1}, {\bf u}; \boldsymbol{\theta}_L)$};
        \node [output, right of=last, node distance=1cm] (output) {};

        \draw [->] (input) -- node[name=u] {\scriptsize ${\bf z}_0$} (first);
        \draw [->] (first) -- node[name=u] {} (dots1);
        \draw [->] (dots1) -- node {\scriptsize ${\bf z}_{l-1}$} (middle);
        \draw [->] (middle) -- node [name=y] {\scriptsize ${\bf z}_l$}(dots2);
         \draw [->] (dots2) -- node [name=y] {}(last);
         \draw [->] (last) -- node[name=u] {\scriptsize ${\bf z}_L$} (output);

    \end{tikzpicture}
    
    \caption{Unrolled network $\boldsymbol{\Phi}({\bf u}; \boldsymbol{\theta})$. Each unrolled layer resembles an update rule $\phi$ of a standard algorithm whose hyperparameters $\boldsymbol{\theta} = \{\boldsymbol{\theta}_l\}_l$ are now set free to learn.}
    \label{fig:unrolled_net}
\end{figure}

The unrolled architecture is trained to learn the parameters $\boldsymbol{\theta} = \{\boldsymbol{\theta}_l \}_{l=1}^L$ that minimize the loss
\begin{equation}\label{eq:optimizer_gen}
    \begin{split}
        \underset{\boldsymbol{\theta}}{\text{argmin}} \quad &  \mathbb{E} \big[\|\boldsymbol{\Phi}({\bf u}; \boldsymbol{\theta}) - {\bf z}^*\|_2^2 \big]\\
        \text{s.t.} \ \quad & {\bf z}^* \in \underset{{\bf z} \in {\cal Z}}{\text{argmin}} \ f({\bf z};{\bf u}), \ \forall {\bf u} \in {\cal U}.
    \end{split}
\end{equation}
The objective is to minimize the distance between the output ${\bf z}_L =: \boldsymbol{\Phi}({\bf u}; \boldsymbol{\theta})$ and a stationary point ${\bf z}^*$, averaged over a dataset of $({\bf u}, {\bf z}^*)$ pairs. The constraints in \eqref{eq:optimizer_gen} are implicit since we resort to a numerical algorithm $\cal A$ to find ${\bf z}^*$ before training the unrolled network. Therefore, \eqref{eq:optimizer_gen} reduces to a typical statistical risk minimization problem.

One advantage of replacing an iterative algorithm with unrolled one is to accelerate the optimizer's convergence. This is not surprising since, during training, we search for the optimal parameters of the standard algorithm that facilitate convergence in a finite, often small, number of steps.

\subsection{Federated Learning}
Our interest, in this paper, is restricted to the case where the objective function $f$ is a learning problem itself.
In particular, we consider federated learning over a network of $n$ nodes that periodically coordinate to train a \emph{single} statistical model $\boldsymbol{\Psi}: {\cal X} \to {\cal Y}$, parameterized by ${\bf w} \in \mathbb{R}^d$, to fit a pair of random variables ${\bf x} \in {\cal X}$ and ${\bf y} \in {\cal Y}$. The nodes communicate over a network, represented by an undirected connected graph ${\cal G} = ({\cal V}, {\cal E})$, where ${\cal V}=\{1, \dots, n\}$ denotes the set of nodes and ${\cal E} \subseteq {\cal V} \times {\cal V}$  denotes the set of edges.
We denote the neighborhood of node $i$ by ${\cal N}_i = \{j \in {\cal V} | (i,j) \in {\cal E}\}$, within which the agent transmits its current estimate of ${\bf w}$. 

The federated learning problem can be cast as the constrained problem
\begin{equation}\label{eq:FedLess}
    \tag{FL}
    \begin{split}
        \underset{{\bf w}_1, \dots, {\bf w}_n }{\min} \quad & f({\bf W}) \coloneqq \frac{1}{n} \sum_{i=1}^n f_i({\bf w}_i),\\
    {\text{s.t.} \quad \quad ~~\: } &
        {\bf w}_i = \frac{1}{|{\cal N}_i|} \sum_{j \in {\cal N}_i} {\bf w}_j, \quad \forall i\in{\cal V},
    \end{split}
\end{equation}
where ${\bf w}_i$ is a \textit{local} version of the global variable ${\bf w}$ stored at agent $i$, and all ${\bf w}_i$'s are arranged in the rows of the matrix ${\bf W} \in \mathbb{R}^{n \times d}$. 
The local objective function $f_i({\bf w}_i) = \mathbb{E} [ \ell(\boldsymbol{\Psi}({\bf x}_i; {\bf w}_i), {\bf y}_i)]$ represents a supervised loss, averaged over a local data distribution ${\cal D}_i$ over the space of data pairs ${\bf x}_i \in {\cal X}$ and ${\bf y}_i \in {\cal Y}$. As in \eqref{eq:optimizer_gen}, $f({\bf W})$ is also a function in the data distribution, which is omitted here for simplicity.

\begin{figure}[t!]
    \centering
    \includegraphics[trim=3.1in 4.1in 3.2in 3.4in, clip, width=.48\textwidth]{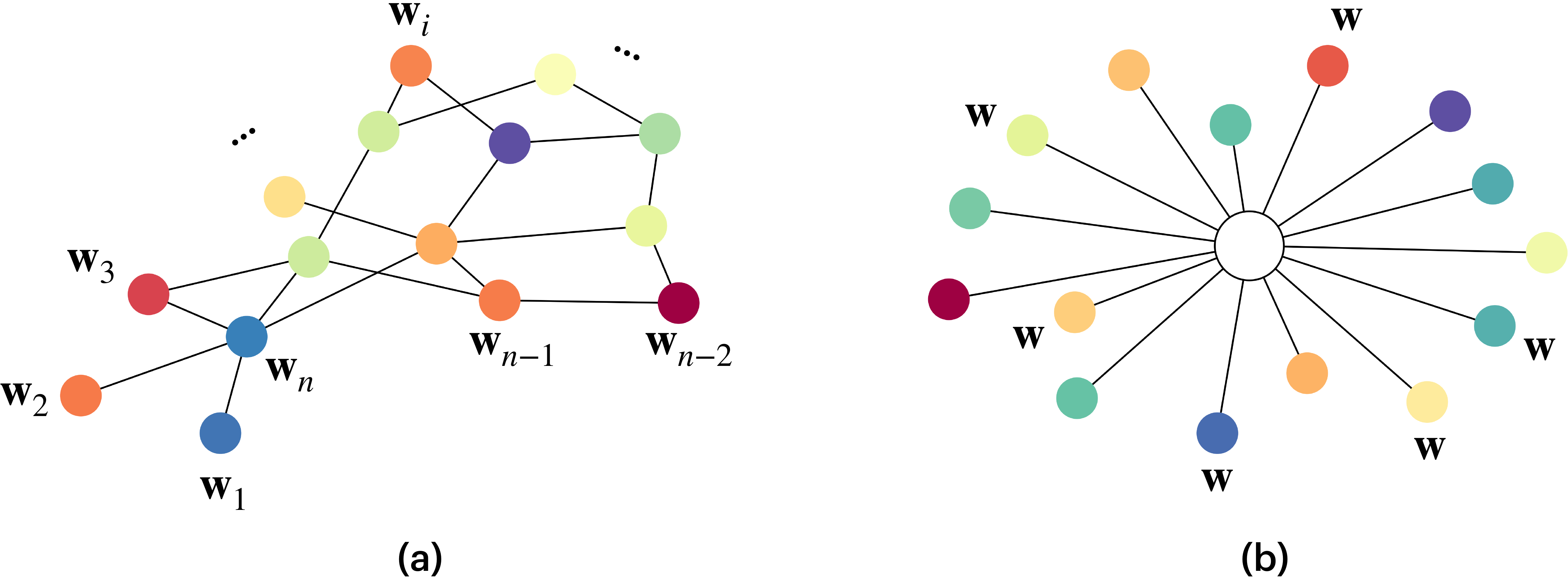}
    \vspace{-.2in}
    \caption{The formulation in~\eqref{eq:FedLess} supports both (a) \emph{decentralized} federated learning, where each agent $i\in\{1, \dots, n\}$ has a local variable ${\bf w}_i$, and (b) \emph{classical} federated learning, where a central server node ensures that all local variables are equal across the network.}
    \label{fig:FL_topology}
\end{figure}

It is evident that, in this formulation of \eqref{eq:FedLess}, each agent learns a separate model ${\bf w}_i$.
The constraints herein, however, mandate that each local variable remains equal to the average of the direct neighbors' local variables. When satisfied, these constraints boil down to constraints of the form ${\bf w}_i = {\bf w}_j$ for all $i$ and $j$ due to the connectivity and symmetry of the graph, hence leading to consensus among agents.

The above formulation generalizes the classical FL problem \citep{McMahan2016} to any graph topology, including cases that do not employ a server node (see Figure~\ref{fig:FL_topology}-a). The following remark shows under which conditions \eqref{eq:FedLess} collapses to the classical form.

\begin{remark} \label{rm:classicalFL}
    The classical FL problem can be restored from \eqref{eq:FedLess} if we choose the underlying graph $\cal G$ to have a star topology, with the center node being a server node (see Figure~\ref{fig:FL_topology}-b). Under this topology, only the server node aggregates weights from all other nodes and subsequently sends the updated weight back to the network. In each iteration, all ${\bf w}_i$'s are then guaranteed to be the same, ruling out the need for consensus constraints.
\end{remark}

\subsection{Unrolling for Federated Learning}
We aim to train an unrolled network $\boldsymbol{\Phi}(\cdot; \boldsymbol{\theta})$ to solve \eqref{eq:FedLess} and find ${\bf W}^*$. The unrolled architecture, similar to \cref{fig:unrolled_net}, unfolds a standard algorithm $\cal A$ in $L$ layers, each of which has the form $ \phi({\bf W}_{l-1}, {\bf D}; \boldsymbol{\theta}_l)$. Here, ${\bf W}_l$ is the output of the $l$-th layer, and ${\bf D}$ is a dataset of data pairs $(\bf{x}, {\bf y})$. The structure of map $\phi$ depends on the standard algorithm being unrolled, which we discuss in detail in Section \ref{sec:unrolling}.

Since the objective function $f$ is a learning problem itself, the training problem \eqref{eq:optimizer_gen}, evaluated over pairs of $({\bf D}, {\bf W}^*)$, becomes a meta-training problem--observe the change of notation from $({\bf u}, {\bf z}^*)$ in \eqref{eq:optimizer_gen} to $({\bf D}, {\bf W}^*)$ here. Each entry in the meta-training dataset is a downstream dataset $\bf D$ accompanied with a model ${\bf W}^*$ that fits the data. However, finding ${\bf W}^*$ for each entry is a computational burden that makes \eqref{eq:optimizer_gen} hard to solve.

To overcome this challenge, we train the unrolled architecture using the unsupervised loss 
\begin{equation}\label{eq:optimizer}
    \begin{split}
        \underset{\boldsymbol{\theta}}{\text{argmin}} \quad &  \mathbb{E} \big[{f}(\boldsymbol{\Phi}({{\bf D}}; \boldsymbol{\theta})) \big].
    \end{split}
\end{equation}
This unsupervised formulation is also a meta-training problem. In fact, \eqref{eq:optimizer} parameterizes the optimization variable $\bf W$ in \eqref{eq:FedLess} with a parameter $\boldsymbol{\theta}$, the unrolled network. Then, instead of searching for ${\bf W}^*$, it finds $\boldsymbol{\theta}^*$ that minimizes the \eqref{eq:FedLess}'s loss function $f$, averaged over a set of FL problems.

Two other challenges, however, become evident. The first one is an artifact of posing the unrolled training as a meta-training problem. In particular, we now need to feed
a whole dataset to each layer of the unrolled network (see \cref{fig:unrolled_net}), which requires the layers to have giant input sizes and is almost impractical.
The other challenge stems from using the loss functions in \eqref{eq:optimizer_gen} and \eqref{eq:optimizer}, which are evaluated at the last layer of the unrolled networks without regularization over the intermediate layers.
As discussed in \cite{hadou2023robust}, without regularization, unrolled networks generate trajectories that do not descend toward the minimum; instead, they learn maps that generate random trajectories and only hit the minimum at the final unrolled layer. In our case, this lack of convergence hinders the generalizability of the unrolled optimizer to query datasets. 
In the following section, we tackle these two issues in our proposed training method, SURF.

\section{
Proposed Method} \label{sec:surf}
To tackle the aforementioned challenges, we introduce Stochastic UnRolled Federated learning, or SURF, a training method for unrolled optimizers that guarantees their convergence. 
SURF resolves the challenge of feeding massive and variable-size datasets using
\emph{stochastic} unrolling, where we feed each layer $l \in \{1, \dots, L\}$ of the unrolled network with a small \emph{fixed-size batch} ${\bf B}_l$ sampled independently and uniformly at random from the dataset $\bf D$. Each layer is then defined as
\begin{equation}
    \tag{U-Layer}
    {\bf W}_l = \phi({\bf W}_{l-1}, {\bf B}_l; \boldsymbol{\theta}_l),
\end{equation}
where the initial estimate ${\bf W}_0$ is drawn from a Gaussian distribution ${\cal N}(\boldsymbol{\mu_0}, \sigma_0^2 {\bf I})$.
It is pertinent to recall that ${\bf W}_l \in \mathbb{R}^{n \times d}$ contains the weight at agent $i$ in its $i$-th row. Similarly, ${\bf B}_l \in \mathbb{R}^{n \times b}$ has the mini-batch at agent $i$ flattened in its $i$-th row. The size of the mini-batch per agent is at least the size of the local dataset divided by $L$ to ensure that all the examples in the downstream dataset has been fed to the unrolled architecture at some layer.

SURF guarantees convergence by imposing \emph{descending constraints} at each unrolling layer. 
The stochastic unrolled federated learning problem can then be formulated as 
\begin{equation}\label{eq:surf}
    \tag{SURF}
    \begin{split}
        \min_{\boldsymbol{\theta}} \quad & \mathbb{E} \big[{f}(\boldsymbol{\Phi}({\bf D}; \boldsymbol{\theta})) \big]\\
        {s. t.} \quad \ &
        \mathbb{E} \big[ \| \nabla f({\bf W}_{l})\|\ - (1-\epsilon) \ \|  \nabla f({\bf W}_{l-1}) \|  \big] \leq 0 ,  \ \forall l,
    \end{split}
\end{equation}
where $\nabla$ denotes stochastic gradients, $\|\cdot\|$ is the Frobenius norm, and $\epsilon  \in (0,1)$ is a design parameter. 
The descending constraints force the gradients to decrease over the layers despite the randomness introduced by relying on a few data points to estimate a descent direction. Intuitively, this would stimulate the unrolled optimizer to converge to a \emph{stationary} point, i.e., ${\bf W}_l \to {\bf W}^*$, on average. Observe that the loss function $f$ is probably non-convex with respect to ${\bf w}_i$ (see \eqref{eq:FedLess}), and therefore, we consider convergence to local minima. A question now arises regarding how to solve the constrained problem \eqref{eq:surf}.


\subsection{Training}
To find the minimizer of \eqref{eq:surf}, we leverage the constrained learning theory (CLT) \citep{chamon2022constrained} by appealing to its dual problem. We formulate the latter by finding the saddle point of the \emph{empirical} Lagrangian function 
\begin{equation}\label{eq:emplagrang}
    \begin{split}
        \widehat{\cal L} & (\boldsymbol{\theta}, {\boldsymbol \lambda}) =  \widehat{\mathbb{E}} \big[{f}(\boldsymbol{\Phi}({\bf D}; \boldsymbol{\theta})) \big] \\ &
        + 
        \sum_{l=1}^L {\lambda}_{l} \widehat{\mathbb{E}} \big[ \| \nabla f({\bf W}_{l})\|\  - (1-\epsilon) \ \|  \nabla f({\bf W}_{l-1}) \|  \big],
    \end{split}
\end{equation}
where ${\boldsymbol \lambda} \in {\mathbb{R}^{L}_+}$ is a vector collecting the dual variables $\lambda_{l}$, and $\widehat{\mathbb{E}}$ denotes the sample mean. The empirical dual problem can then be cast as 
\begin{equation}\label{eq:dual}
    \begin{split}
        \widehat{D}^* = \max_{{\boldsymbol \lambda} \in \mathbb{R}^{L}_+} \ \min_{\boldsymbol{\theta}} \ \widehat{\cal L}(\boldsymbol{\theta}, {\boldsymbol \lambda}).
    \end{split}
\end{equation}
Equation \eqref{eq:dual} is an unconstrained optimization problem that can be solved by alternating between minimizing the Lagrangian with respect to $\boldsymbol{\theta}$ for a fixed $\boldsymbol{\lambda}$ and then maximizing over the latter, as described in Algorithm \ref{alg:PD}. 

Nevertheless, \eqref{eq:dual} is not equivalent to \eqref{eq:surf} due to the non-convexity of the problem among other factors. CLT, however, analyzes the gap between the two problems and shows that a solution to the former is a near-optimal and near-feasible solution to the latter.
\begin{theorem}[CLT (informal)]\label{thm:CLTinf}
A stationary point of \eqref{eq:dual} is a near-optimal and near-feasible solution to \eqref{eq:surf}  under some mild assumptions. That is, for each $l$,
\begin{equation}\label{eq:constraints_inf}
    \begin{split}
         \mathbb{E} \big[\| \nabla f({\bf W}_{l})\|\  - (1-\epsilon) \ \|  \nabla f({\bf W}_{l-1}) \|  \big] \leq \zeta(Q, \delta),
    \end{split}
\end{equation}
with probability $1-\delta$, and $\zeta(Q, \delta)$ measures the sample complexity.
\end{theorem}
\begin{algorithm}[t]
\caption{Primal-Dual Meta-Training }\label{alg:PD}
\begin{algorithmic}
 \STATE {\bfseries Input:} Meta-training dataset $\bar{\bf D} = \{{\bf D}_q \}_{q=1}^{Q}$.
 \STATE Initialize $\boldsymbol{\theta} =\{\boldsymbol{\theta}_l\}_{l=1}^L$ and ${\boldsymbol \lambda} = \{ {\boldsymbol \lambda}_{l} \}_{l=1}^L$.
\FOR{each epoch}
    \FOR{each batch}
    \STATE Sample a dataset from $\bar{\bf D}$ and compute $\widehat{\cal L}(\boldsymbol{\theta}, {\boldsymbol \lambda})$ as in \cref{fig:unrolled_FL}.
    \FOR{$l = 1, \dots, L$}
    \STATE Update variables at layer $l$:
    \begin{align}
        \boldsymbol{\theta}_l &\gets [\boldsymbol{\theta}_l - \mu_{\theta} \nabla_{\boldsymbol{\theta}_l} \widehat{\cal L}(\boldsymbol{\theta}, {\boldsymbol \lambda})], \label{eq:ADAM}\\
        {\boldsymbol \lambda}_{l} & \gets [{\boldsymbol \lambda}_{l} + \mu_{\lambda} \nabla_{{\boldsymbol \lambda}_{l}} \widehat{\cal L}(\boldsymbol{\theta}, {\boldsymbol \lambda})]_+.\label{eq:lr_dual}
    \end{align}
 \STATE {\bfseries Return:} $\boldsymbol{\theta}_l^* \gets \boldsymbol{\theta}_l, \forall l\in\{1, \dots, L\}$.
 \ENDFOR
 \ENDFOR
 \ENDFOR
\end{algorithmic}
\end{algorithm}

 A formal statement of this theorem can be found in Appendix \ref{app:CLT}.
The first implication drawn from Theorem \ref{thm:CLTinf} is that solving \eqref{eq:surf} is equally easy (or hard) as solving its unconstrained alternative in \eqref{eq:optimizer} since both have the same sample complexity. The second implication is that an unrolled optimizer trained via Algorithm \ref{alg:PD} is a probably near-optimal, near-feasible solution to \eqref{eq:surf}. Consequently, each unrolled layer in the trained optimizer satisfies the descending constraints and takes a step in a descent direction with probability $1-\delta$, where $\delta$ depends on the size of the meta-training dataset. Hence, the trained unrolled optimizer can be considered as a stochastic descent algorithm.
\tikzstyle{block} = [draw, fill=lightgray, rectangle, 
    minimum height=1em, minimum width=7em]
\tikzstyle{PSIblock} = [draw, fill=pink, rectangle, 
    minimum height=4em, minimum width=5em]
\tikzstyle{dots} = [rectangle, 
    minimum height=1em, minimum width=1em]
\tikzstyle{highlight} = [draw, rectangle, 
    minimum height=10em, minimum width=7em]
\tikzstyle{sum} = [draw, fill=white, circle, node distance=1cm]
\tikzstyle{input} = [coordinate]
\tikzstyle{output} = [coordinate]
\tikzstyle{tmp} = [coordinate]
\tikzstyle{pinstyle} = [pin edge={to-,thin,black}]
\tikzstyle{pinstyle2} = [pin edge={-to,thin,black}]
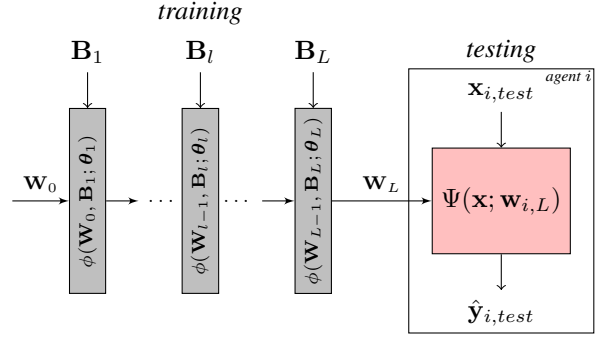
\begin{figure}[t]
    \centering
    \begin{tikzpicture}[auto, node distance=2cm,>=latex']

        \node [input, name=input] {};
        \node [block, right of=input, rotate=90, pin={[pinstyle]right:${{\bf B}_1}$},
                node distance=1cm] (first) {\scriptsize$ \phi({\bf W}_{0}, {{\bf B}_1}; \boldsymbol{\theta}_1)$};
        \node [dots, right of=first,
                node distance=1cm] (dots1) {\scriptsize $\dots$};
        \node [block, right of=dots1, rotate=90, pin={[pinstyle]right:${{\bf B}_l}$},
                node distance=0.5cm] (middle) {\scriptsize $\phi({\bf W}_{l-1}, {{\bf B}_l}; \boldsymbol{\theta}_l)$};
        \node [dots, right of=middle,
                node distance=0.5cm] (dots2) {\scriptsize $\dots$};
        \node [block, right of=dots2, rotate=90, pin={[pinstyle]right:${{\bf B}_L}$},
                node distance=1cm] (last) {\scriptsize $\phi({\bf W}_{L-1}, {{\bf B}_L}; \boldsymbol{\theta}_L)$};
        \node [output, right of=last, node distance=1.5cm] (output) {};
        \node [PSIblock, right of=output, pin={[pinstyle]above:${{\bf x}_{i, test} }$},
                pin={[pinstyle2]below:$\hat{\bf y}_{i, test}$},
                node distance=1cm] (Psi) {$\Psi({\bf x}; {\bf w}_{i,L})$};
        \node [dots, above of=middle, node distance=2.5cm] (text1) {\emph{training}};
        \node [dots, above of=Psi, node distance=2cm] (text2) {\emph{testing}};
        \node [highlight, above of=Psi, node distance=0cm] (highlight1) {};
        \node [tmp, right of = Psi, node distance=0.9cm] (temp1) {};
        \node [dots, above of=temp1, node distance=1.65cm] (text3) {\emph{\tiny agent $i$}};

        \draw [->] (input) -- node[name=u] {\scriptsize ${\bf W}_0$} (first);
        \draw [->] (first) -- node[name=u] {} (dots1);
         \draw [->] (dots2) -- node [name=y] {}(last);
         \draw [->] (last) -- node[name=u] {\scriptsize ${\bf W}_L$} (Psi);

    \end{tikzpicture}
    
    \caption{One iteration of \cref{alg:PD}. One dataset $\bf D$ is chosen randomly from the meta-training dataset and divided into training and testing examples. Mini-batches of the training examples are randomly selected and fed to the unrolled layers (in gray) to predict ${\bf W}_L$. Given the latter, the loss function $\widehat{\cal L}$ is computed over the testing examples, averaged over all agents, and its gradients update the parameters $\boldsymbol{\theta}$ and $\boldsymbol{\lambda}$. }
    \label{fig:unrolled_FL}
\end{figure}


\subsection{Convergence Guarantees}
Despite the above result, finding $\boldsymbol{\theta}^*$ does not directly guarantee its capability to generate a sequence of layers' outputs $\{{\bf W}_l\}_{l=1}^L$ that converges to the optimal solution of \eqref{eq:FedLess}. This is because this convergence requires (almost) all the descending constraints to be satisfied, which has a decreasing probability $(1-\delta)^L$ with the number of layers $L$ despite the fact that these constraints are statistically independent. In Theorem \ref{thm:convergence}, we prove that the trained unrolled optimizer indeed converges to a near-optimal region infinitely often.

\begin{theorem}\label{thm:convergence} For an $M$-Lipschitz loss function $f({\bf W})$ and a sequence of $\{ {\bf W}_l| l\geq 0\}$ that satisfies \cref{thm:CLTinf}, it holds that 
    \begin{equation}
    \lim_{l \rightarrow \infty} \mathbb{E} \Big[ \min_{k\leq l} \|{\nabla} f({\bf W}_{k})\| \Big] \leq \frac{1}{\epsilon} \left(\zeta(Q, \delta) + \frac{\delta M}{1-\delta}\right) \quad a.s.
    \end{equation}
\end{theorem}
The proof constructs a stochastic process $\alpha_l$ that keeps track of the gradient norm until it drops below $\frac{1}{\epsilon} \big(\zeta(Q, \delta) + \frac{\delta M}{1-\delta}\big)$ and shows that $\alpha_l$ converges almost-surely using the supermartingale convergence theorem \citep{robbins_convergence_1971}. The detailed proof of Theorem \ref{thm:convergence} is relegated to Appendix~\ref{app:thm1}. 

The above result implies that the sequence $\{ {\bf W}_l| l\geq 0\}$ infinitely often visit a region around the optimal where the norm of the gradient drops below $\frac{1}{\epsilon} \big(\zeta(Q, \delta) + \frac{\delta M}{1-\delta}\big)$, on average. The size of this near-optimal region depends on the sample complexity of the model $\boldsymbol{\Phi}$, the Lipschitz constant of the loss function and its gradient, and lastly a design parameter $\epsilon$ of the imposed constraints. The larger $\epsilon$, which is equivalent to imposing an aggressive reduction on the gradients (see \eqref{eq:constraints_inf}), the closer we are guaranteed to get to a local optimal ${\bf W}^*$.


In addition to the asymptotic analysis, we aspire to 
characterize an upper bound for the gradient norm after a finite number of layers $L$ in Theorem \ref{thm:rate}.
\begin{theorem}\label{thm:rate}
For a trained unrolled optimizer $\boldsymbol{\theta}^*$ that satisfies Theorem \ref{thm:CLTinf}, the gradient norm achieved after $L$ layers satisfies
\vspace{-5pt}
\begin{equation}
\begin{split}
     \mathbb{E} \big[ {\|{\nabla} f({\bf W}_{L})\|} \big] 
    \leq & (1-\delta)^L(1-{\epsilon})^L \ \mathbb{E} {\|  {\nabla} f({\bf W}_{0})\|} \\ & + \frac{1}{\epsilon} \left( \zeta(Q, \delta) + \frac{\delta M}{1-\delta} \right).
\end{split}
\end{equation}
\end{theorem}
The proof is relegated to Appendix \ref{app:thm3}. The theorem shows that the unrolled optimizer trained via SURF has an exponential rate of convergence. Our method \emph{surpasses} the sublinear convergence rates of the state-of-the art FL methods, e.g., SGD, decentralized FedAvg \citep{sun2022decentralized}, FedProx \citep{li2020federated}, MOON \citep{Li_2021_CVPR} among others. What sets our method apart is the fact that the unrolled optimizer learns to navigate the optimization landscape, unrestrained to the gradient direction, thus enabling it to discover descending directions that facilitate quick convergence.

\section{GNN-based Unrolled DGD}\label{sec:unrolling}
As could be envisaged, SURF, as a training method, is agnostic to the iterative update map $\phi$ that we choose to unroll. However, the standard update rule and the unrolled architecture should accommodate the requirements of the \eqref{eq:FedLess} problem, which are i) to permit distributed execution and ii) to satisfy the consensus constraints of \eqref{eq:FedLess}. In this section, we pick DGD as an example of a  decentralized algorithm that satisfies the two requirements and unroll it using GNNs, which also can be executed distributedly.

DGD is a distributed iterative algorithm that relies on limited communication between agents. At each iteration $l$, the updating rule $\phi$ of DGD has the form
\begin{equation}\label{eq:DGD}
    {\bf w}_{i}(l) = \sum_{j \in {\cal N}_i \cup \{i\}} \alpha_{ij} {\bf w}_{j}(l-1) - \beta \nabla f_i({\bf w}_{i}(l-1)), \forall i,
\end{equation}
where $f_i$ is the local objective function, $\beta$ is a fixed step size and $\alpha_{ij} = \alpha_{ji}$. The weights $\alpha_{ij}$ are chosen such that  $\sum_{j=1}^n \alpha_{ij} = 1$ for all $i$ to ensure that \eqref{eq:DGD} converges \citep{nedic_distributed_2009}. The update rule in \eqref{eq:DGD} can be interpreted as letting the agents descend in the opposite direction of the local gradient $\nabla f_i({\bf w}_{i}(l-1))$ as they move away from the (weighted) average of their neighbors' estimates ${\bf w}_{j}(l-1)$. 
Each iteration can then be divided into two steps; first the agents aggregate information from their direct neighbors and then they update their weights based on the gradient of their local objective functions. 

Unrolling \eqref{eq:DGD} could yield different unrolled architectures according to which parameters we choose to unroll. In our proposal, we utilize graph filters and single-layer fully-connected perceptrons, as we show in \cref{fig:UDGD} and the following subsections. In addition,
we differentiate between two cases: i) decentralized FL over an arbitrary graph with no servers, and ii) classical FL over a star graph.

\subsection{U-DGD for Decentralized FL}
We unfold the update rule of DGD in a learnable neural layer of the form
\begin{equation}\label{eq:U-DGD}
    \tag{U-DGD}
    \begin{split}
        {\bf w}_{i,l} & =: [\phi({\bf W}_{l-1}, {{\bf B}_l}; {\bf h}_l, {\bf M}_l, {\bf d}_l)]_i \\
        &= [{\bf H}_{l}({\bf W}_{l-1})]_i - \sigma \left( {\bf M}_l \left[{\bf w}_{i, l-1} \Vert  {\bf b}_{i,l} \right]   + {\bf d}_l\right),
    \end{split}
\end{equation}
where $[.]_i$ refers to the $i$-th row of a matrix, $\Vert$ denotes a vertical vector concatenation, ${\bf b}_{i,l} = [{\bf B}_l]_i$ is the mini-batch used by agent $i$ at layer $l$, and $\sigma$ is a non-linear activation function. In \eqref{eq:U-DGD}, we replace the first term in \eqref{eq:DGD} with a learnable graph filter (see \eqref{eq:GF}) and the second term with a single fully-connected perceptron. \cref{fig:UDGD} illustrates these operations executed at agent $i$, which represents the interior of one unrolled block in \cref{fig:unrolled_FL}. The learnable parameters of $\phi$ are then $\boldsymbol{\theta} = \{ {\bf h}_l, {\bf M}_l, {\bf d}_l \}$, which we elaborate on in the rest of this subsection.
\tikzstyle{block} = [draw, rectangle, 
    minimum height=2.5em, minimum width=4em]
\tikzstyle{sum} = [draw, circle]
\tikzstyle{dots} = [rectangle, 
    minimum height=1em, minimum width=1em]
\tikzstyle{Giantblock} = [draw, rectangle, thick,
    minimum height=9em, minimum width=14em]
\tikzstyle{Giantblock_tr} = [draw, rectangle, thick,
    minimum height=9em, minimum width=14em, fill=lightgray, opacity=0.3]
\tikzstyle{sum} = [draw, fill=white, circle, node distance=1cm]
\tikzstyle{input} = [coordinate]
\tikzstyle{output} = [coordinate]
\tikzstyle{tmp} = [coordinate]
\tikzstyle{pinstyle} = [pin edge={to-,thin,black}]
\tikzstyle{pinstyle2} = [pin edge={-to,thin,black}]
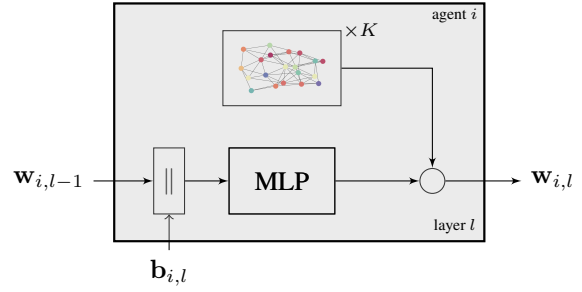
\begin{figure}[t]
    \centering
    \begin{tikzpicture}[auto, node distance=2cm,>=latex']

        \node [block, node distance=1cm] (GF) {\includegraphics[width=3.8em]{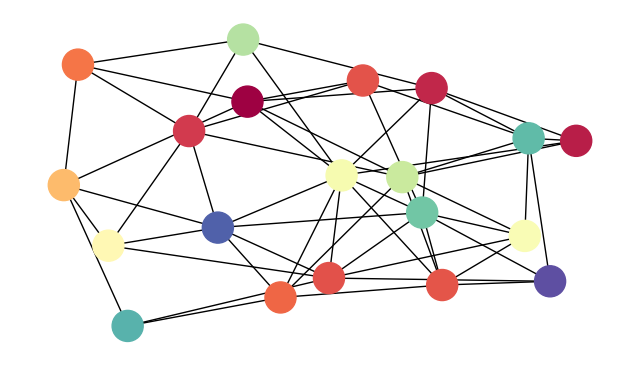}};
        \node [block, below of = GF, node distance=1.5cm] (MLP) {MLP};
        \node [sum, right of = MLP, node distance = 2cm] (sum) {};
        \node [block, left of = MLP, minimum width=1em, node distance = 1.5cm, pin={[pinstyle]below:${\bf b}_{i,l}$}] (con) {$\Vert$};
        \node [Giantblock_tr, fit = (con)(MLP)(GF)(sum), node distance=0cm] (outer) {};
         \node [dots, left of=con, node distance=1.6cm] (first) {${\bf w}_{i, l-1}$};
        \node [dots, right of=sum, node distance=1.6cm] (last) {${\bf w}_{i, l}$};
        \node [tmp, above of = sum, node distance=1.5cm] (temp1) {};
        \node [Giantblock, fit = (con)(MLP)(GF)(sum), node distance=0cm] (outer) {};
        \node [block, below of = GF, node distance=1.5cm] (MLP) {MLP};
        \node [tmp, right of = sum, node distance=0.3cm] (temp2) {};
        \node [dots, above of=temp2, node distance=2.2cm] (text) {\tiny agent $i$};
        \node [dots, below of=temp2, node distance=0.6cm] (text2) {\tiny layer $l$};
        \node [tmp, right of = MLP, node distance=1.05cm] (temp3) {};
        \node [dots, above of=temp3, node distance=2cm] (K) {\scriptsize $\times K$};

        \draw [->] (first) -- node[name=u] {} (con);
        \draw [->] (con) -- node[name=u] {} (MLP);
        \draw [->] (GF) -- (temp1) -| node[name=u] {} (sum);
         \draw [->] (MLP) --  node[name=u] {} (sum);
         \draw [->] (sum) -- node[name=u] {} (last);

    \end{tikzpicture}
    \vspace{-10pt}
    \caption{An unrolled layer $\phi({\bf W}_{l-1}, {{\bf B}_l}; {\bf h}_l, {\bf M}_l, {\bf d}_l)$ of U-DGD at agent $i$. The block on top is a graph filter, parameterized by ${\bf h}_l$, which performs $K$ communication rounds, and the block underneath is a single-layer MLP, parameterized by ${\bf M}_l$ and ${\bf d}_l$. All agents share the same learnable parameters.}
    \label{fig:UDGD}
\end{figure}

The graph filter, the building block of GNNs \citep{gama_graphs_2020}, aggregates information from up to $K$-hop neighbors,
\begin{equation}\label{eq:GF}
    {\bf H}({\bf W}_{l-1}) = \sum_{k=0}^{K} h_{k,l} {\bf S}^k {\bf W}_{l-1},
\end{equation}
where $\bf S$ is a graph shift operator, e.g., graph adjacency or Laplacian. The graph filter executes a linear combination of information gathered from up to $K$-hop neighbors, and, in turn, requires $K$ communication rounds. Here, the filter coefficients ${\bf h}_l = \{h_{k,l}\}_{k=0}^{K}$ that weigh the information aggregated from different hop neighbors are the learnable parameters. Equation \eqref{eq:GF} and the first term of \eqref{eq:DGD} are essentially the same when $K$ is set to $1$ and $h_k$ to $1$ for all $k$ and $\bf S$ is chosen to be the (normalized) graph adjacency matrix. In U-DGD, however, the goal is to learn the weights $h_{k,l}$ to accelerate the unrolled network's convergence.

The other component of \eqref{eq:U-DGD} is a single-layer fully-connected perceptron, which is implemented locally and whose weights ${\bf M}_l \in \mathbb{R}^{n \times d+b}$ and ${\bf d}_l \in \mathbb{R}^{d+b}$ are shared among all the agents. The role of this neural perceptron is to perform the local update of the weights once every $K$ communication rounds. The input to this perceptron at each agent is the previous estimate ${\bf w}_{i, l-1} \in \mathbb{R}^d$ concatenated with a mini-batch ${\bf b}_{i,l} \in \mathbb{R}^b$. Each batch is a concatenation of the sampled data points, where the input data and label of one example follow each other. 

\begin{remark}
Since the parameters of the fully-connected perceptron are shared between all the agents, U-DGD learners inherit the permutation equivariance of graph filters and graph neural networks, as well as transferability to graphs with different sizes \citep{Luana_graphon} and stability to small graph perturbations \citep{gama19, gama20, hadouspace, hadou2023space}.
\end{remark}

\subsection{U-DGD for Classical FL}
As per \cref{rm:classicalFL}, classical FL can be retrieved from \eqref{eq:FedLess} when the underlying graph is a star topology. In such a topology, the nodes only communicate with one another through a central node, which can function as a server in classical FL settings. Therefore, U-DGD can extend to classical FL with a minor adjustment to account for the fact that the server does not possess local data.

The unrolled layer of the server, denoted as node $0$, employs only a graph filter to aggregate information from the rest of the network, i.e.,
\begin{equation}\label{eq:server}
    \begin{split}
        {\bf w}_{0,l} =  h_{l} [{\bf S}]_0 {\bf W}_{l-1}.
    \end{split}
\end{equation}
The fully-connected perceptron in \eqref{eq:U-DGD} is excluded here since the server node does not locally execute a weight update. Furthermore, \eqref{eq:server} is identical to the first row of \eqref{eq:GF} when $K$ is $1$ and $k=0$ is omitted. This means that the server only performs a weighted average of the information received from the other nodes based on the values at the first row of $\bf S$.
The rest of the nodes with indices $i>0$, on the other hand, apply the exact form of \eqref{eq:U-DGD} while constraining $K$ to $1$.

\section{Numerical Experiments}\label{sec:results}
In this section, we run experiments to show the convergence rate of U-DGD networks, trained via SURF, under different settings.

\textbf{Set-up.} We consider a network of $n=100$ agents, which collaborate to train a softmax layer of an image classifier. The softmax layer is fed by the outputs of the convolutional layers of a ResNet18 backbone, whose weights are pre-trained and kept frozen during the training process. 
To train a U-DGD optimizer via SURF, we consider a meta-training dataset, which consists of $600$ class-imbalanced datasets. Each dataset has a different label distribution and contains $6,000$ images (that is, $45$ training examples/agent and $15$ for testing) that are evenly divided between the agents.

\textbf{Meta-training}. At each epoch, we randomly choose one image dataset from the meta-training dataset and feed its $45$ training examples/agent to a $10$-layer U-DGD network in mini-batches of $10$ examples/agent at each layer (see \cref{fig:unrolled_FL}). The training loss is computed over the $10$ testing examples/agent and optimized using ADAM with a learning rate $\mu_{\theta} = 10^{-2}$ and a dual learning rate $\mu_{\lambda} = 10^{-2}$. We utilize ReLU activation functions at each layer, and the constraint parameter $\epsilon$ is set to $0.01$. The performance of the trained U-DGD is examined over a meta-testing dataset that consists of $30$ class imbalanced datasets, each of which also has $45$ training examples and $15$ for testing per agent. Similar to training, the training examples are fed to the U-DGD in mini-batches while the testing examples are used to compute the testing accuracy. The results are reported for CIFAR10 dataset \citep{krizhevsky2009learning}. All experiments were run on an NVIDIA\textsuperscript{\tiny \textregistered} GeForce RTX\texttrademark~3090 GPU.\footnote{Our code is available at: \href{https://github.com/SMRhadou/fed-SURF}{https://github.com/SMRhadou/fed-SURF}.}

\begin{figure*}[t]
    \centering    \includegraphics[width=0.31\textwidth]{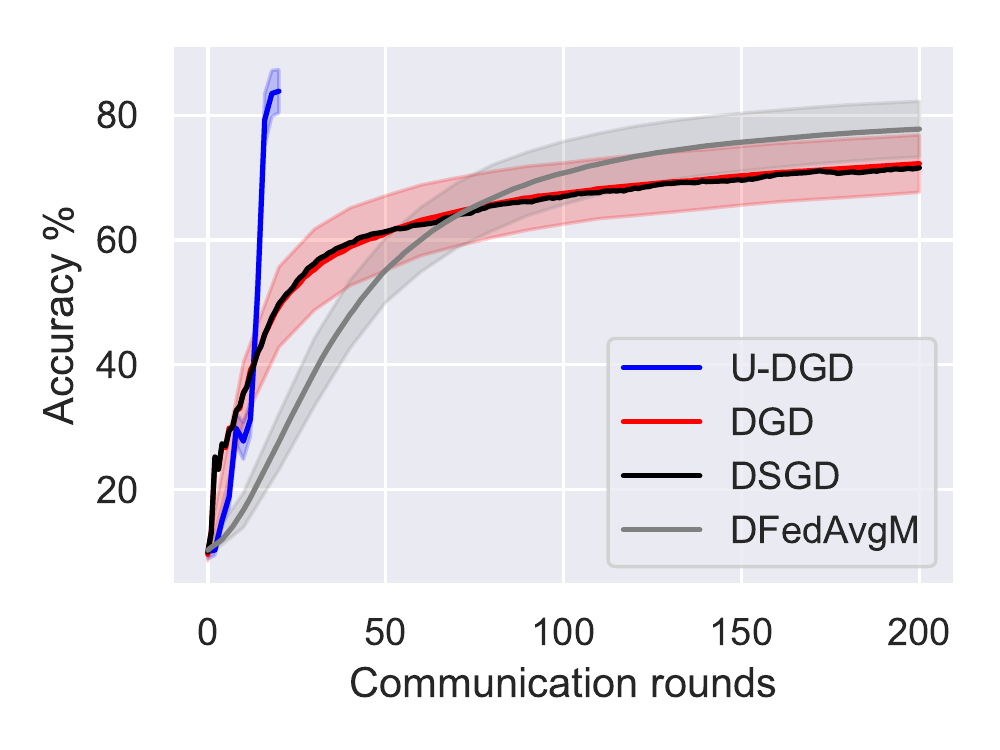}
     \includegraphics[width=0.31\textwidth]{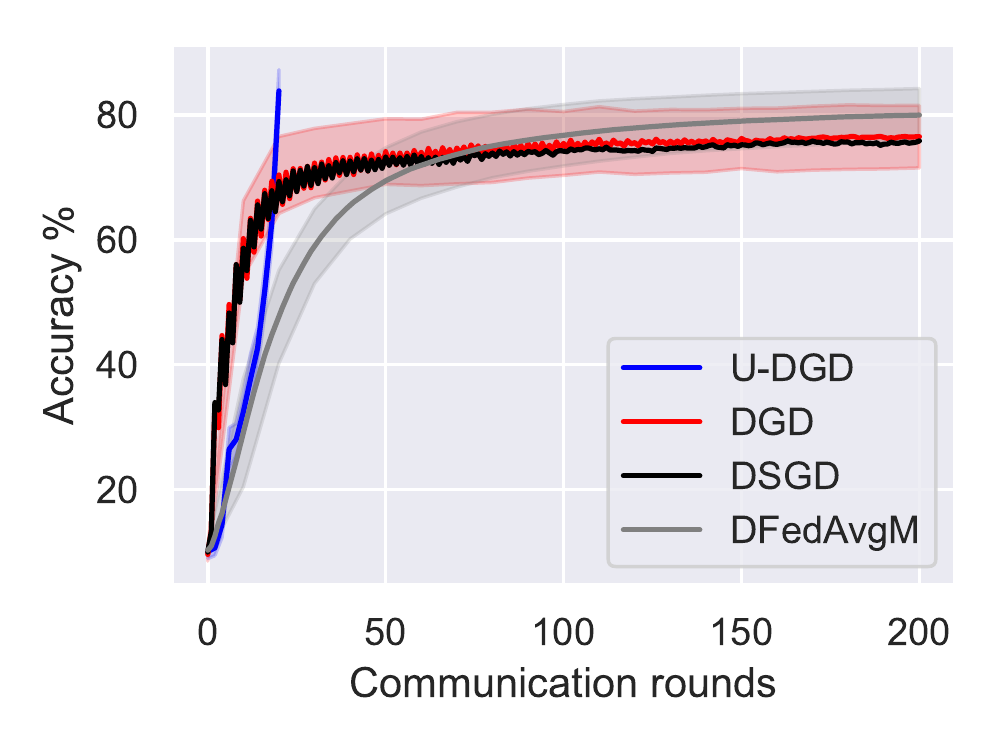}
     \includegraphics[width=0.31\textwidth]{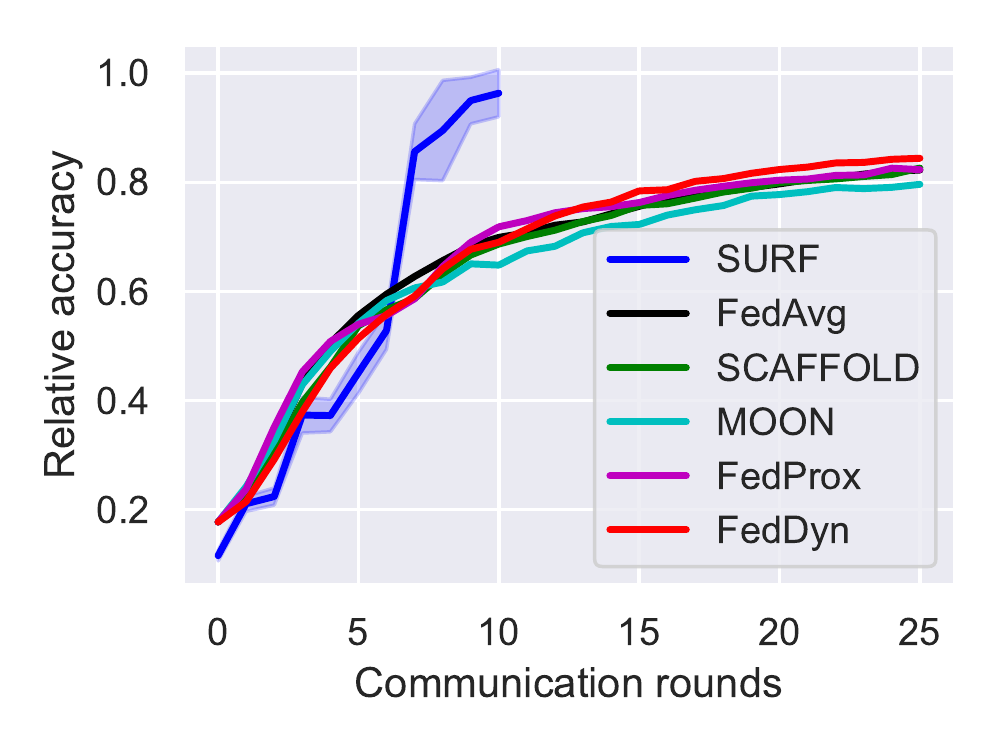}
    \caption{\textbf{Convergence rate.} Comparisons between the accuracy of U-DGD and state-of-the-art FL methods for both i) decentralized FL over $3$-degree regular graphs (left) and random graphs (middle), and ii) classical FL with a star graph (right). U-DGD scores higher convergence rates in all settings surpassing both decentralized and centralized FL methods.
    }
    \label{fig:speed}
\end{figure*}

\begin{figure*}[t]
    \centering    \includegraphics[width=0.31\textwidth]{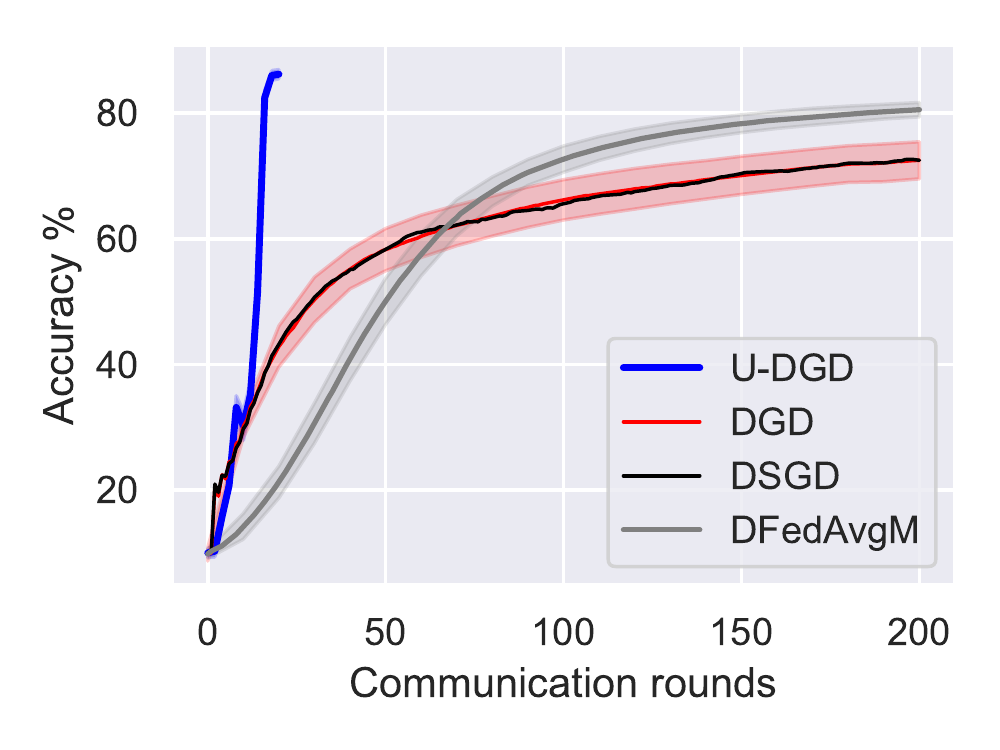}
     \includegraphics[width=0.31\textwidth]{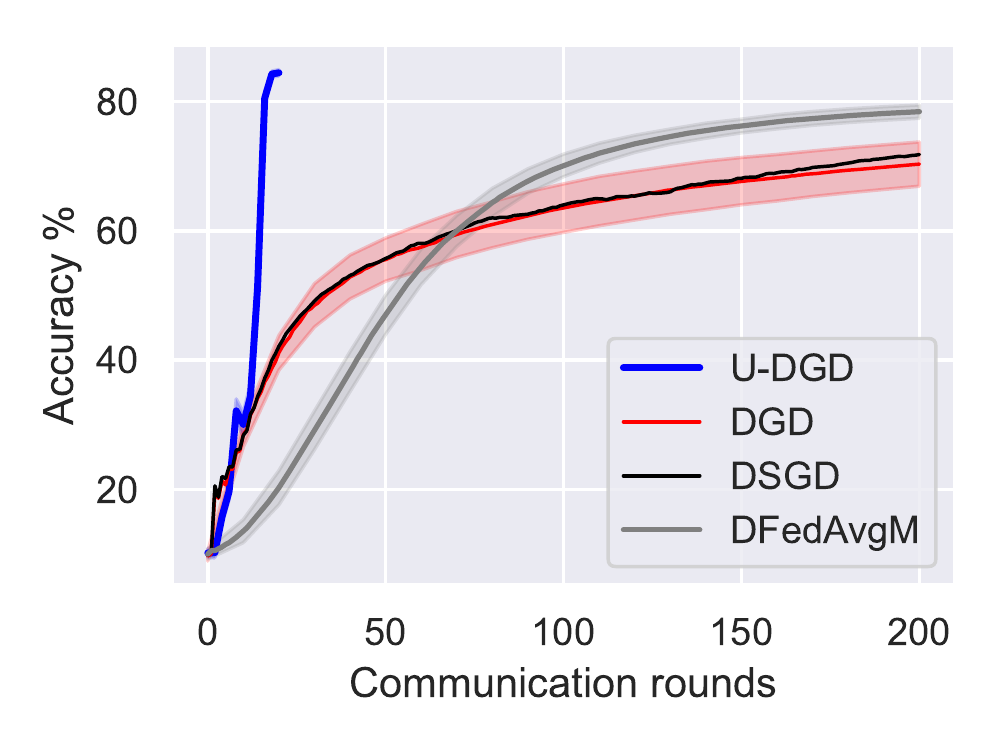}
     \includegraphics[width=0.31\textwidth]{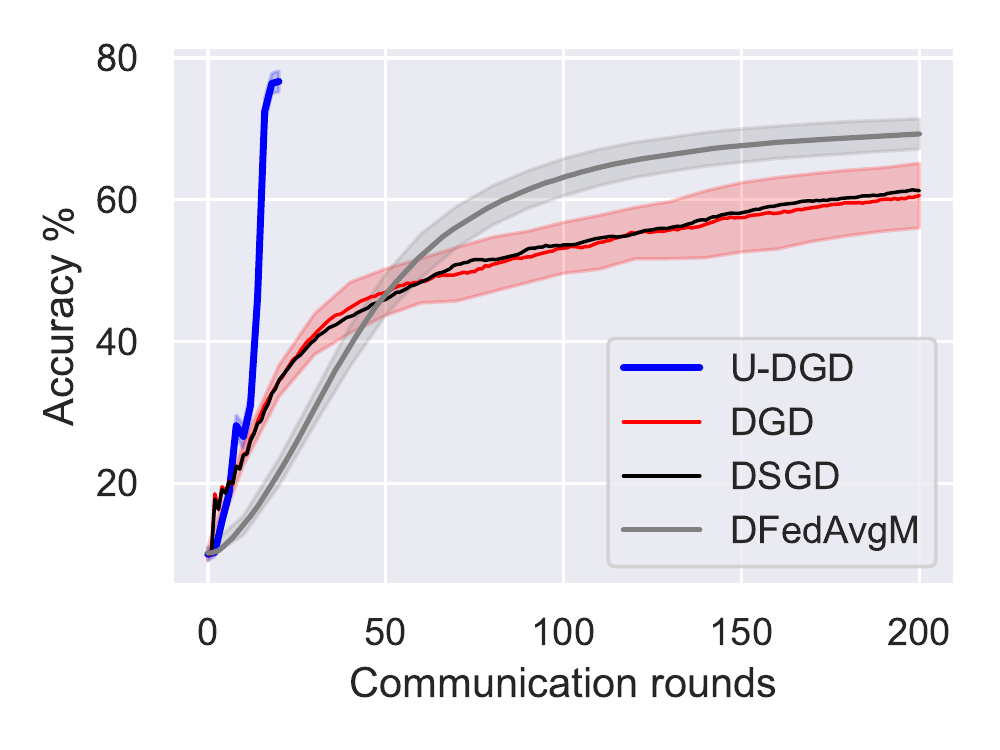}
    \caption{\textbf{Heterogeneous settings.} Comparisons between the accuracy of U-DGD and other decentralized benchmarks, evaluated over $30$ class-imbalanced CIFAR10 datasets sampled according to a Dirichlet distribution with a concentration parameter (Left) $\alpha =1$, (Middle) $\alpha =0.7$, and (Right) $\alpha =0.3$. The lower $\alpha$ is, the more heterogeneous the agents are. U-DGD is more robust than the other benchmarks.}
    \label{fig:drift}
\end{figure*}

\textbf{Decentralized FL over arbitrary graphs.} We consider two graph topologies, specifically $3$-degree regular graphs and random graphs, to create decentralized FL environments. In the former topology, each node connects to exactly $3$ other nodes, while in the latter, an edge is connected between any two nodes with probability $p=0.1$. For both scenarios,
we train a U-DGD that consists of $10$ unrolled layers, each of which employs a graph filter that aggregates information from up to two neighbors (i.e., $K=2$). This results in a total of $20$ communication rounds between the agents. 

We compare the accuracy of U-DGD to other decentralized FL benchmarks: DGD (c.f. \eqref{eq:DGD}), distributed stochastic gradient descent (DSGD), and decentralized federated averaging (DFedAvgM) \citep{sun2022decentralized}.
Figure \ref{fig:speed} shows the convergence rates of all methods. U-DGD exhibts a faster convergence rate as it takes only $20$ communication rounds to achieve performance higher than that achieved by the others in $200$ communication rounds.

\textbf{Classical FL over a star graph}. 
In addition, we consider a star graph with one of the nodes acting as a server.
We set $\mu_\theta = 10^{-3}$, $\epsilon = 0.1$, and $K = 1$ while the rest of the parameters are kept the same as they were in the decentralized experiment. 
We compare the convergence rate of U-DGD with other FL benchmarks: FedAvg \citep{McMahan2016}, SCAFFOLD \citep{karimireddy2020scaffold}, MOON \citep{Li_2021_CVPR}, FedProx \citep{li2020federated}, and FedDyn \citep{acar2021federated}. For fair comparisons, all the methods, including U-DGD, let only $10$ agents to share their updated weights with the server node at each communication round.
Figure \ref{fig:speed} shows that U-DGD has a notably faster convergence than all the benchmarks.
The figure also shows the relative accuracy of these methods compared to centralized training. It suggests that SURF almost achieves the same performance attained by central training in 10 communication rounds while the other benchmarks need 25 rounds to reach almost $80\%$ of the centralized performance. 

\textbf{Heterogeneous settings.}
We evaluate our unrolled model, U-DGD, on a network of heterogeneous agents that sample their data according to a Dirichlet distribution with a concentration parameter $\alpha$. A lower value of $\alpha$ indicates greater heterogeneity among the agents.  The network connecting the agents is structured as a $3$-degree regular graph. Figure \ref{fig:drift} provides the accuracy averaged over $30$ heterogeneous downstream datasets for different values of $\alpha$. The plots reveal that heterogeneity does not significantly affect the convergence rate of U-DGD. In fact, U-DGD consistently outperforms all other benchmarks in terms of accuracy. Additionally, it is observed that the performance of all methods deteriorates when $\alpha$ decreases. However, the degradation in the performance of U-DGD is comparatively slower when compared to the other methods. 

An ablation study of the descending constraints is deferred to Appendix \ref{app:ExtraExp} due to the limited space.

\section{Conclusions}
In this paper, we proposed a new framework, called SURF, that introduces stochastic algorithm unrolling to federated learning scenarios. 
To showcase the merits of SURF, we unrolled DGD in a GNN-based unrolled architecture that can solve both decentralized and classical FL. The main takeaway is that U-DGD, trained using SURF, achieves convergence rates faster than the sublinear rates of the state-of-the-art FL methods.
The convergence of the unrolled architecture is guaranteed by the imposition of descending constraints over the unrolled layers during training.

There are several directions for future work. One possible avenue is to explore other standard distributed algorithms to design unrolled architectures for personalized federated learning scenarios.
Moreover, privacy is a critical concern in federated learning, since even though the agents do not share their data, they communicate their evaluated gradients, which can be exploited in inferring the data. Unrolled optimizers are prone to the same privacy issues since the input of the fully-connected perceptron can be inferred from its outputs \citep{Fredrikson15}. Methods inspired by differential privacy~\citep{Abadi15, arachchige2019local} and secure aggregation~\citep{so2021securing, elkordy2022much} can be further explored in the context of unrolling. 



\bibliography{Bib
}
\bibliographystyle{icml2024}

\newpage
\appendix
\onecolumn

\section{Extended Related Work}\label{app:related}
\textbf{Learning to Optimize/Learn (L2O/L2L).} Our work is mostly related to the broad research area of L2O \citep{chen_learning_2021}, which aims to automate the design of optimization methods by training optimizers on a set of training problems. L2O has achieved notable success in many optimization problems including $\ell_1$-regularization \citep{gregor10}, neural-network training \citep{Andrychowicz16, Ravi2016OptimizationAA}, minimax optimization \citep{shen2021learning}, and black-box optimization \citep{chen17} among many others.

Prior work in L2O can be divided into two categories; model-free and model-based optimizers. Model-free L2O aims to train an iterative update rule that does not take any analytical form and relies mainly on general-purpose recurrent neural network (RNNs) and long short-term memory networks (LSTMs) \citep{Andrychowicz16, chen17, Lyu2017LearningGD, wichrowska17, Xiong20, Jiang21}. Model-based L2O, on the other hand, provides compact, interpretable learning networks by taking advantage of both model-based algorithms and data-driven learning paradigms \citep{gregor10, Greenfeld19}. 
As part of this category, algorithm unrolling aims to unroll the hyperparameters of a standard iterative algorithm in a neural network to learn them. The seminal work \citep{gregor10} unrolled iterative shrinkage thresholding algorithm (ISTA) for sparse coding problems. Following \citep{gregor10}, many other algorithms have been unrolled, including, but not limited to, projected gradient descent \citep{Giryes18}, the primal-dual hybrid gradient algorithm \citep{Jiu2020ADP, Cheng19}, and Frank-Wolfe \citep{liu2019frank}.

Learning to learn (L2L) refers to frameworks that extend L2O to training neural networks in small data regimes, e.g., few-shot learning \citep{Triantafillou2020}. Learning to learn has strong ties to meta-learning, but they differ in their ultimate goal; meta-learning, e.g., model-agnostic meta-learning (MAML) \citep{finn_model-agnostic_2017}, aims to learn an initial model that can be fine-tuned in a few gradient updates, whereas L2L aims to learn the gradient update and the step size.
General purpose LSTM-based models, e.g., \citep{Ravi2016OptimizationAA, Andrychowicz16, Li2017MetaSGDLT} are the most popular among L2L models. 

\textbf{Algorithm Unrolling using GNNs.}
Algorithm unrolling has also been introduced to distributed optimization problems with the help of graph neural networks (GNNs). One of the first distributed algorithms to be unrolled was weighted minimum mean-square error
(WMMSE)~\citep{shi2011iteratively}, which benefited many applications including wireless resource allocation \citep{Chowdhury21, Li22} and multi-user multiple-input multiple-output (MU-MIMO) communications \citep{Hu21, zhou22, Ma22, Pellaco22, Schynol22, Schynol23}. Several other distributed unrolled networks have been developed for graph topology inference \citep{pu2021learning}, graph signal denoising \citep{Chen21denoising, Nagahama21} and computer vision \citep{lin2022ru}, among many others. 

\textbf{Decentralized Federated Learning.} There have been many efforts in recent years to enable federated learning without the aid of a server, e.g., \citep{kalra2023decentralized, sun2022decentralized, wang2022peer, tedeschini2022decentralized, ye2022decentralized, wink2021approach} to name a few. These efforts have benefited from the advances in decentralized algorithms, such as decentralized SGD \citep{pmlr-v119-koloskova20a, wang2021cooperative}, asynchronous decentralized SGD \citep{lian2018asynchronous}, and alternating direction method of multipliers (ADMM) \citep{wei2012distributed, shi2014linear}.  
Our proposed method deviates from these studies in that we use a meta approach to learn the optimizer instead of using state-of-the art optimizers.

\section{Constrained Learning Theory}\label{app:CLT}
In this section, we provide a rigorous statement for CLT theorem and the assumptions under which it holds.
\begin{assumption}\label{A1}
The loss function $f(\cdot)$ in \eqref{eq:surf} and the gradient norm $\| {\nabla} f(\cdot)\|$ are both bounded and $M$-Lipschitz continuous functions.
\end{assumption}
\begin{assumption}\label{A2}
Let $\widehat{\mathbb{E}}$ be the sample mean evaluated over $Q$ realizations. Then
there exists $\zeta(Q, \delta)\geq 0$ that is monotonically decreasing with $Q$, for which it holds with probability $1-\delta$ that 
\vspace{-5pt}
\begin{enumerate}
    \item $|\mathbb{E}[f(\boldsymbol{\Phi}({\boldsymbol{\vartheta}}; \boldsymbol{\theta}))] - \widehat{\mathbb{E}}[f(\boldsymbol{\Phi}({\boldsymbol{\vartheta}}; \boldsymbol{\theta}))]| \leq \zeta(Q, \delta)$, and
    \item $|\mathbb{E}[ \| \nabla f({\bf W}_l({\boldsymbol{\vartheta}}; \boldsymbol{\theta})) \| ] - \widehat{\mathbb{E}}[ \| \nabla f({\bf W}_l({\boldsymbol{\vartheta}}; \boldsymbol{\theta})) \| ]| \leq \zeta(Q, \delta)$ for all $l$ and all $\boldsymbol{\theta} \in \mathbb{R}^p$.
\end{enumerate}
\end{assumption}
\begin{assumption}\label{A3}
Let $\phi_l {\scriptscriptstyle\circ} \dots {\scriptscriptstyle\circ} \phi_1 \in {\cal P}_l$ be a composition of $l$ unrolled layers parameterized by $\boldsymbol{\theta}_{1:l}$ and $\overline{\cal P}_l = \overline{conv}({\cal P}_l)$ be the convex hull of ${\cal P}_l$.
Then, for each $\overline\phi_l {\scriptscriptstyle\circ} \dots {\scriptscriptstyle\circ} \overline\phi_1 \in \overline{\cal P}$ and $\nu > 0$, there exists $\boldsymbol{\theta}_{1:l}$ such that $\mathbb{E} \left[ |\phi_l {\scriptscriptstyle\circ} \dots {\scriptscriptstyle\circ} \phi_1({\bf W}_0,\boldsymbol{\vartheta}; \boldsymbol{\theta}_{1:l}) - \overline\phi_l {\scriptscriptstyle\circ} \dots {\scriptscriptstyle\circ} \overline\phi_1({\bf W}_0, \boldsymbol{\vartheta})| \right] \leq \nu$ for all $l$. 
\end{assumption}
\begin{assumption}\label{A4}
There exists $\boldsymbol{\Phi} \in {\cal H}$ that is strictly feasible, i.e., $\mathbb{E} \big[\| \nabla f({\bf W}_{l})\|\  - (1-\epsilon) \ \|  \nabla f({\bf W}_{l-1}) \| \big] < -M\nu, \forall l$, with $M$ and $\nu$ as in Assumptions \ref{A1} and \ref{A3}.
\end{assumption}
\begin{assumption}\label{A5}
The conditional distribution $p(\boldsymbol{\vartheta}|{\bf W})$ is non-atomic for all ${\bf W}$.
\end{assumption}

The above assumptions can be easily satisfied in practice. Assumption \ref{A1} requires the loss function and its gradient to be smooth and bounded. Assumption \ref{A2} identifies the sample complexity, which is a common assumption when handling statistical models. Moreover, Assumption \ref{A3}
forces the parameterization $\boldsymbol{\theta}_l$ to be sufficiently rich at each layer $l$, which is guaranteed by modern deep learning models.
Assumption \ref{A4} ensures that the problem is feasible and well posed, which is guaranteed since \eqref{eq:surf} mimics the parameters of a standard iterative solution. Finally, Assumption \ref{A5} can be satisfied using data augmentation.
\begin{theorem}[CLT \citep{chamon2022constrained}]\label{thm:CLT}
Let $P^*$ be the optimal value of \eqref{eq:surf} and $(\boldsymbol{\theta}^*, \boldsymbol{\lambda}^*)$ be a stationary point of \eqref{eq:dual}. Under Assumptions \ref{A1}- \ref{A5}, it holds, for some constant $\rho$, that 
\begin{equation}\label{eq:nearOptimal}
    |P^* - \widehat{D}^*| \leq M \nu + \rho \ \zeta(Q, \delta), \ \text{and}
\end{equation} 
\begin{equation}\label{eq:constraints}
    \begin{split}
         \mathbb{E} \big[\| \nabla f({\bf W}_{l})\|\  - (1-\epsilon) \ \|  \nabla f({\bf W}_{l-1}) \|  \big] \leq \zeta(Q, \delta), \quad  \forall l,
    \end{split}
\end{equation}
with probability $1-\delta$ each and with $\rho \geq \max \{ \| \boldsymbol{\lambda}^*\|, \| \overline{\boldsymbol{\lambda}}^* \| \}$, where $\overline{\boldsymbol{\lambda}}^* = \textup{argmax}_{{\boldsymbol \lambda}} \ \min_{\boldsymbol{\theta}} \ {\cal L}(\boldsymbol{\theta}, {\boldsymbol \lambda})$.
\end{theorem}

CLT asserts that the gap between the two problems is affected by a smoothness constant $M$, the richness of the parameterization $\boldsymbol{\theta}$, and the sample complexity.

\section{Proofs}
In this section, we provide the proofs for our theoretical results after introducing the following notation.
Consider a probability space $(\Omega, {\cal F}, P)$, where $\Omega$ is a sample space, ${\cal F}$ is a sigma algebra, and $P:{\cal F} \rightarrow [0,1]$ is a probability measure. With a slight abuse of this measure-theoretic notation, we write $P(X=0)$ instead of $P(\{ \omega: X(\omega) = 0\})$, where $X: \Omega \rightarrow \mathbb{R}$ is a random variable, to keep equations concise. We define a filtration of $\cal F$ as $\{{\cal F}_l\}_{l>0}$, which can be thought of as an increasing sequence of $\sigma$-algebras with ${\cal F}_{l-1} \subset {\cal F}_l$. We assume that the outputs of the unrolled layers ${\bf W}_l$ are adapted to ${\cal F}_l$, i.e., ${\bf W}_l \in {\cal F}_l$. Intuitively, the filtration ${\cal F}_l$ describes the information at our disposal at step $l$, which includes the outputs of each layer up to layer $l$, along with the initial estimate ${\bf W}_0$.

In our proofs, we use a supermartingale argument, which is commonly used to prove the convergence of stochastic descent algorithms. A stochastic process $X_k$ is said to form a supermartingale if $\mathbb{E}[X_k | X_{k-1}, \dots, X_0] \leq X_{k-1}$. This inequality implies that given the past history of the process, the future value $X_k$ is not, on average, larger than the latest one. With this definition in mind, we provide the proof of \cref{thm:convergence}.

\subsection{Proof of Theorem \ref{thm:convergence}}\label{app:thm1}
This proof follows the lines of the proof of Theorem 2 in \cite{hadou2023robust}.

Let $A_l \in {\cal F}_l$ be the event that the constraint \eqref{eq:constraints} at layer $l$ is satisfied. By the law of total expectation, we have 
\begin{equation}\label{eq:totalexp}
\begin{split}
    \mathbb{E} \big[ {\|{\nabla} f({\bf W}_{l})\|} \  \big]  = 
    P(A_l)\mathbb{E} \Big[ {\|{\nabla} f({\bf W}_{l})\|} \ | A_l \Big]
    + P(A_l^c)\mathbb{E} \Big[ {\|{\nabla} f({\bf W}_{l})\|}  \ | A_l^c \Big],
\end{split}
\end{equation}
with $P(A_l)=1-\delta$.
On the right-hand side, the first term represents the conditional expectation when the constraint is satisfied and, in turn, is bounded above according to \eqref{eq:constraints}. The second term is concerned with the complementary event $A_l^c \in {\cal F}_l$, when the constraint is violated. The conditional expectation in this case can also be bounded since i) the gradient norm $\| {\nabla} f({\bf W}_{l})\| \leq M$ for all ${\bf W}_{l}$ since $f$ is $M$-Lipschitz according to Assumption \ref{A1}, and ii) the expectation of a random variable cannot exceed its maximum value, i.e, $\mathbb{E}\| {\nabla} f({\bf W}_{l})\| \leq \max_{{\bf W}_l}\| {\nabla} f({\bf W}_{l})\| \leq M$ by Cauchy-Schwarz inequality.  Substituting in \eqref{eq:totalexp} results in an upper bound of
\begin{equation}\label{eq:bound1}
\begin{split}
    \mathbb{E} \big[ {\|{\nabla} f({\bf W}_{l})\|} \big] 
    & \leq (1-\delta)(1-{\epsilon}) \ \mathbb{E} {\|  {\nabla} f({\bf W}_{l-1})\|} + (1-\delta)\zeta(Q, \delta) + \delta M,
\end{split}
\end{equation}
almost surely.

In the rest of the proof, we leverage the supermartingale convergence theorem to show that \eqref{eq:bound1} indeed implies the required convergence. We start by defining a sequence of random variables $\{Z_l\}_l$ each of which has a degenerative distribution such that
\begin{equation}
    {Z_l} = {\mathbb{E}\|  {\nabla} f({\bf W}_{l})\|} \quad a.s. \quad \forall l.
\end{equation}
Then, we form  a supermartingale-like inequality using the law of total expectation. That is, we have
\begin{equation}\label{eq:bound2}
\begin{split}
    \mathbb{E} [Z_l | \ {\cal F}_{l-1} ]
    & \leq (1-\delta)(1-{\epsilon}) \  Z_{l-1} + (1-\delta)\zeta(Q, \delta) + \delta M\\
    & = (1-\delta) \  Z_{l-1} - (1-\delta) \Big( \epsilon Z_{l-1} - \zeta(Q, \delta) - \frac{\delta M}{1-\delta} \Big).
\end{split}
\end{equation}
The structure of the proof is then divided into two steps. First, we prove that when $l$ grows, $Z_l$ almost surely and infinitely often achieves values below $\frac{1}{\epsilon} \big(\zeta(Q, \delta) + \delta M/1-\delta\big)$.  Second, we show that this is also true for the gradient norm $\|  {\nabla} f({\bf W}_{l})\|$ itself. This implies that the outputs of the unrolled layers enter a near-optimal region infinitely often.

To tackle the first objective, we define the lowest gradient norm achieved, on average, up to layer $l$ as $Z_l^\text{best} = \min_{k\leq l} \{Z_k \}$. To ensure that $Z_l$ enters this region infinitely often, it suffices to show that
\begin{equation}\label{eq:goal1}
    \lim_{l \rightarrow \infty}  Z_l^\text{best} \leq \frac{1}{\epsilon} \big(\zeta(Q, \delta) + \delta M/1-\delta\big) \quad a.s.
\end{equation}
To show that the above inequality is true, we start by defining the sequences
\begin{equation}\label{eq:sequences}
    \begin{split}
        \alpha_l & :=  Z_l \cdot \mathbf{1} \Big\{  Z_l^\text{best} > \frac{1}{\epsilon} \big(\zeta(Q, \delta) + \delta M/1-\delta\big) \Big\},\\
        \beta_l & := \Big( \epsilon Z_{l} - \zeta(Q, \delta) - \frac{\delta M}{1-\delta} \Big) \mathbf{1}\Big\{  Z_l^\text{best} > \frac{1}{\epsilon} \big(\zeta(Q, \delta) + \delta M/1-\delta\big) \Big\},
    \end{split}
\end{equation}
where $\mathbf{1}\{.\}$ is an indicator function.
The first sequence $\alpha_l$ tracks the values of $Z_l$ until the best value $Z_l^\text{best}$ drops below the threshold $\frac{1}{\epsilon} \big(\zeta(Q, \delta) + \delta M/1-\delta\big)$ for the first time. After this point, the best value stays below the threshold since $Z_{l+1}^\text{best} \leq Z_l^\text{best}$ by definition, which implies that the indicator function stays zero and $\alpha_l = 0$. In other words, we have
\begin{equation}\label{eq:alpha_l}
    \alpha_l = \left\{ \begin{array}{cc}
        Z_l & l < T \\
        0 & \text{otherwise},
    \end{array}\right.
\end{equation}
with $T := \min \{l \ | \ Z_l^\text{best} \leq \frac{1}{\epsilon} \big(\zeta(Q, \delta) + \delta M/1-\delta\big) \}$. Similarly, the sequence $\beta_l$ follows the values of $\epsilon Z_{l} - \zeta(Q, \delta) - \frac{\delta}{1-\delta} M$ until it falls below zero for the first time, which implies that $\beta_l \geq 0$ by construction. Moreover, it also holds that $\alpha_l \geq 0$  for all $l$ since $Z_l$ is always non-negative.

We now aim to show that $\alpha_l$ forms a supermartingale, so we can use the supermartingale convergence theorem to prove \eqref{eq:goal1}. This requires finding an upper bound of the conditional expectation $\mathbb{E}[\alpha_l | {\cal F}_{l-1}]$. We separate this expectation into two cases, $\alpha_{l-1}=0$ and $\alpha_{l-1} \neq 0$, and use the law of total expectation to write
\begin{equation} \label{eq:SM}
    \mathbb{E}[\alpha_l | {\cal F}_{l-1}] = \mathbb{E}[\alpha_l | {\cal F}_{l-1}, \alpha_{l-1}=0] P(\alpha_{l-1}=0)
    + \mathbb{E}[\alpha_l | {\cal F}_{l-1}, \alpha_{l-1} \neq 0] P(\alpha_{l-1}\neq 0).
\end{equation}
First, we focus on the case when $\alpha_{l-1}=0$, and for conciseness, let $\eta := \frac{1}{\epsilon} \big(\zeta(Q, \delta) + \delta M/(1-\delta\big))$ be the radius of the near-optimal region centered around the optimal. Equation \eqref{eq:sequences} then implies that the indicator function is zero, i.e., $Z_l^\text{best} \leq \eta$, since the non-negative random variable $Z_l$ cannot be zero without $Z_l^\text{best} \leq \eta$. It also follows that $\beta_{l-1}$ is zero since it employs the same indicator function as $\alpha_l$. As we discussed earlier, once $\alpha_{l-1} = 0$, all the values that follow are also zero, i.e., $\alpha_{k}=0, \ \forall k\geq l-1$ (c.f. \eqref{eq:alpha_l}). Hence, the conditional expectation of $\alpha_l$ can be written as
\begin{equation}\label{eq:SM_part1}
    \mathbb{E}[\alpha_l | {\cal F}_{l-1}, \alpha_{l-1}=0] = 0 =: (1-\delta)(\alpha_{l-1} - \beta_{l-1}).
\end{equation}
On the other hand, when $\alpha_{l-1}\neq 0$, the conditional expectation follows from the definition in \eqref{eq:sequences},
\begin{equation} \label{eq:SM_part2}
    \begin{split}
        \mathbb{E}[\alpha_l | {\cal F}_{l-1}, \alpha_{l-1} \neq 0] & = \mathbb{E}[Z_l \cdot \mathbf{1}\{  Z_l^\text{best} > \eta \} | {\cal F}_{l-1}, \alpha_{l-1} \neq 0]\\
        & \leq \mathbb{E}[Z_l | {\cal F}_{l-1}, \alpha_{l-1} \neq 0]\\
        & \leq (1-\delta) \  Z_{l-1} - (1-\delta) \Big( \epsilon Z_{l-1} - \zeta(Q, \delta) - \frac{\delta M}{1-\delta} \Big)\\
        & = (1-\delta) (\alpha_{l-1} - \beta_{l-1}).
    \end{split}
\end{equation}
The first inequality is true since the indicator function is at most one, and the second inequality is a direct application of \eqref{eq:bound2}. The last equality results from the fact that the indicator function $\mathbf{1}\{ Z_l^\text{best} > \eta \}$ is 1 since $\alpha_{l-1} \neq 0$, which implies that $\alpha_{l-1} = Z_{l-1}$ and $\beta_{l-1} = \epsilon Z_{l-1} - \zeta(Q, \delta) - \frac{\delta}{1-\delta} M$. Combining the results in \eqref{eq:SM_part1} and \eqref{eq:SM_part2} and substituting in \eqref{eq:SM}, it finally follows that 
\begin{equation}\label{eq:final_SM}
\begin{split}
    \mathbb{E}[\alpha_l | {\cal F}_{l-1}] & \leq (1-\delta) (\alpha_{l-1} - \beta_{l-1}) [P(\alpha_{l-1}= 0) + P(\alpha_{l-1}\neq 0)]\\
    & = (1-\delta) (\alpha_{l-1} - \beta_{l-1}),
\end{split}
\end{equation}
and we emphasize that both $\alpha_{l-1}$ and $\beta_{l-1}$ are non-negative by definition.

Given \eqref{eq:final_SM}, it follows from supermartingale convergence theorem \citep[Theorem 1]{robbins_convergence_1971} that (i) $\alpha_l$ converges almost surely, and (ii) $\sum_{l=1}^\infty \beta_l$ is almost surely summable (i.e., finite). When the latter is written explicitly, we get
\begin{equation}\label{eq:finiteSum}
    \sum_{l=1}^\infty \Big( \epsilon Z_{l} - \zeta(Q, \delta) - \frac{\delta M}{1-\delta} \Big) \mathbf{1}\{  Z_l^\text{best} > \eta \} < \infty, \quad a.s.,
\end{equation}
The almost sure convergence of the above sequence implies that the limit inferior and limit superior coincide and 
\begin{equation}\label{eq:liminfTotal}
    \liminf_{l \rightarrow \infty} \Big( \epsilon Z_{l} - \zeta(Q, \delta) - \frac{\delta M}{1-\delta} \Big) \mathbf{1}\{  Z_l^\text{best} > \eta \} = 0, \quad a.s.
\end{equation}
The latter is true if either there exist a sufficiently large $l$ such that $Z_l^\text{best} \leq \eta = \frac{1}{\epsilon} \big(\zeta(Q, \delta) + \delta M/1-\delta\big)$ or it holds that
\begin{equation}\label{eq:liminf}
     \liminf_{l \rightarrow \infty}  \Big( \epsilon Z_{l} - \zeta(Q, \delta) - \frac{\delta M}{1-\delta} \Big) = 0, \quad a.s.
\end{equation}
The above equation can be re-written as $\sup_{l} \inf_{m\geq l}  Z_{m} = \frac{1}{\epsilon} \big(\zeta(Q, \delta) + \frac{\delta M}{1-\delta} \big)$. Hence, there exists some large $l$ where $Z_l^\text{best} \leq \sup_{l} \inf_{m\geq l}  Z_{m}$, which again reaches the same conclusion. This proves the correctness of \eqref{eq:goal1}.

To this end, we have shown the convergence of $Z_l^\text{best}$, which was defined as the best \textit{expected} value of $\| {\nabla} f({\bf W}_{l})\|$. It is still left to show the convergence of the random variable $\| {\nabla} f({\bf W}_{l})\|$ itself. Start with writing $Z_l = \int \| {\nabla} f({\bf W}_{l})\| dP$, which turns \eqref{eq:liminf} into
\begin{equation}
    \liminf_{l \rightarrow \infty}  \int \epsilon \| {\nabla} f({\bf W}_{l})\|  dP = \zeta(Q, \delta) + \frac{\delta M}{1-\delta}, \quad a.s.
\end{equation}
By Fatou's lemma \citep[Theorem 1.5.5]{durrett2019probability}, it follows that 
\begin{equation}
       \int \liminf_{l \rightarrow \infty} \epsilon \| {\nabla} f({\bf W}_{l})\| dP \leq \liminf_{l \rightarrow \infty}  \int \epsilon \| {\nabla} f({\bf W}_{l})\| dP = \zeta(Q, \delta) + \frac{\delta M}{1-\delta}.
\end{equation}
We can bound the left hand side from below by defining $f^\text{best}_l := \min_{k\leq l} \| {\nabla} f({\bf W}_{k})\|$ as the lowest gradient norm achieved up to layer $l$. By definition, $f^\text{best}_l \leq \liminf_{l \rightarrow \infty} \| {\nabla} f({\bf W}_{l})\|$ for sufficiently large $l$. Therefore, we get
\begin{equation}
        \epsilon \int  f^\text{best}_l dP \leq \epsilon \int \liminf_{l \rightarrow \infty}  \| {\nabla} f({\bf W}_{l})\|  dP \leq  \zeta(Q, \delta) + \frac{\delta M}{1-\delta}, \quad a.s.
\end{equation}
for some large $l$. Equivalently, we can write that
\begin{equation}\label{eq:res1}
        \lim_{l \rightarrow \infty} \int  f^\text{best}_l dP \leq \frac{1}{\epsilon} \left( \zeta(Q, \delta) + \frac{\delta M}{1-\delta} \right), \quad a.s.
\end{equation}
which completes the proof.


\subsection{Proof of Theorem \ref{thm:rate}}\label{app:thm3}
This proof follows the lines of the proof of Lemma 1 in \cite{hadou2023robust}. 
\begin{proof}
First, we recursively unroll the right-hand side of \eqref{eq:bound1} to evaluate the reduction in the gradient norm $\mathbb{E} {\|{\nabla} f({\bf W}_{l})\|}$ after $l$ layers. This leads to the inequality
\begin{equation}\label{eq:bound_recursive}
\begin{split}
    \mathbb{E} \big[ {\|{\nabla} f({\bf W}_{l})\|} \big] 
    \leq \ \ & (1-\delta)^l(1-{\epsilon})^l \ \mathbb{E} {\|  {\nabla} f({\bf W}_{0})\|}\\
    & + \sum_{i=0}^{l-1} (1-\delta)^{i-1}(1-{\epsilon})^{i-1} \Big[ (1-\delta)\zeta(Q, \delta) + \delta M \Big].
\end{split}
\end{equation}
The right-hand side can be further simplified by evaluating the geometric sum resulting in
\begin{equation}\label{eq:bound_recursive2}
\begin{split}
    \mathbb{E} \big[ {\|{\nabla} f({\bf W}_{l})\|} \big] 
    \leq \ \ & (1-\delta)^l(1-{\epsilon})^l \ \mathbb{E} {\|  {\nabla} f({\bf W}_{0})\|}\\
    & +\frac{1 - (1-\delta)^l(1-{\epsilon})^l}{1-(1-\delta)(1-{\epsilon})} \Big[ (1-\delta)\zeta(Q, \delta) + \delta M \Big].
\end{split}
\end{equation}

Second, we evaluate the distance between $\mathbb{E} {\|{\nabla} f({\bf W}_{L})\|}$ at the $L$-th layer and its optimal value
\begin{equation}\label{eq:convergence}
\begin{split}
    \Big| \mathbb{E} & \big[ {\|{\nabla} f({\bf W}_{L})\|} \big] - \mathbb{E} \big[ {\|{\nabla} f({\bf W}^*)\|} \big] \Big| \\
    & = \lim_{l \to \infty} \Big| \mathbb{E} \big[ {\|{\nabla} f({\bf W}_{L})\|} \big] 
    - \mathbb{E} [ \min_{k\leq l} \|{\nabla} f({\bf W}_{k})\| ]
    + \mathbb{E} [ \min_{k\leq l} \|{\nabla} f({\bf W}_{k})\| ]
    -  \mathbb{E} \big[ {\|{\nabla} f({\bf W}^*)\|} \big] \Big|.
\end{split}
\end{equation}
We add and subtract $\mathbb{E} [ \min_{k\leq l} \|{\nabla} f({\bf W}_{k})\| ]$ in the right-hand side while imposing the limit when $l$ goes to infinity so we can use triangle inequality. We, hence, get
\begin{equation}\label{eq:convergence2}
\begin{split}
    \Big| \mathbb{E} \big[ {\|{\nabla} f({\bf W}_{L})\|} \big] - \mathbb{E} & \big[ {\|{\nabla} f({\bf W}^*)  \|} \big]\Big| \\
    & \leq \lim_{l \to \infty} \Big| \mathbb{E} \big[ {\|{\nabla} f({\bf W}_{L})\|} \big] 
    - \mathbb{E} [ \min_{k\leq l} \|{\nabla} f({\bf W}_{k})\| ] \Big| \\
    & \quad +  \lim_{l \to \infty} \Big| \mathbb{E} [ \min_{k\leq l} \|{\nabla} f({\bf W}_{k})\| ]
    -  \mathbb{E} \big[ {\|{\nabla} f({\bf W}^*)\|} \big] \Big|.
\end{split}
\end{equation}
Note that the gradient of $f$ at the stationary point ${\bf W}^*$ is zero. Therefore, the second term on the right-hand side is upper bounded according to Theorem \ref{thm:convergence}. 

The final step required to prove Theorem \ref{thm:rate} is to evaluate the first term in \eqref{eq:convergence2}. To do so, we observe that 
\begin{equation}\label{eq:convergence3}
\begin{split}
    \lim_{l \to \infty} \Big| \mathbb{E} \big[ {\|{\nabla} f({\bf W}_{L})\|} \big] 
    - \mathbb{E} [ \min_{k\leq l} \|{\nabla} f({\bf W}_{k})\| ] \Big| = 
     \mathbb{E} \big[ {\|{\nabla} f({\bf W}_{L})\|} \big] 
    - \lim_{l \to \infty} \mathbb{E} [ \min_{k\leq l} \|{\nabla} f({\bf W}_{k})\| ].
\end{split}
\end{equation}
This is the case since $ \mathbb{E} \big[ {\|{\nabla} f({\bf W}_{L})\|} \big] $ cannot go below the minimum of the gradient norm when $l$ goes to infinity. Therefore, we can using \eqref{eq:bound_recursive2}
\begin{equation}\label{eq:convergence4}
    \begin{split}
        \lim_{l \to \infty} \Big| & \mathbb{E} \big[ {\|{\nabla} f({\bf W}_{L})\|} \big] 
    - \mathbb{E} [ \min_{k\leq l} \|{\nabla} f({\bf W}_{k})\| ] \Big| \\ & =
    (1-\delta)^L(1-{\epsilon})^L \ \mathbb{E} {\|  {\nabla} f({\bf W}_{0})\|} 
    +\frac{1 - (1-\delta)^L(1-{\epsilon})^L}{1-(1-\delta)(1-{\epsilon})} \Big[ (1-\delta)\zeta(Q, \delta) + \delta M \Big] \\
    & - \lim_{l \to \infty} (1-\delta)^l(1-{\epsilon})^l \ \mathbb{E} {\|  {\nabla} f({\bf W}_{0})\|} 
     - \lim_{l \to \infty} \frac{1 - (1-\delta)^l(1-{\epsilon})^l}{1-(1-\delta)(1-{\epsilon})} \Big[ (1-\delta)\zeta(Q, \delta) + \delta M \Big] \\
     & = (1-\delta)^L(1-{\epsilon})^L \ \mathbb{E} {\|  {\nabla} f({\bf W}_{0})\|} 
    - \frac{(1-\delta)^L(1-{\epsilon})^L}{1-(1-\delta)(1-{\epsilon})} \Big[ (1-\delta)\zeta(Q, \delta) + \delta M \Big] \\
    & \leq (1-\delta)^L(1-{\epsilon})^L \ \mathbb{E} {\|  {\nabla} f({\bf W}_{0})\|}.
    \end{split}
\end{equation}
Note that the first limit in the left-hand side of \eqref{eq:convergence4} goes to zero and the second limit is evaluated as the constant $(1-\delta)\zeta(Q, \delta) + \delta M$ divided by $1-(1-\delta)(1-{\epsilon})$. The final inequality follows since the second term in the second-to-last line is always non-negative.

Combining the two results, we can bound the quantity in \eqref{eq:convergence2} as follows;
\begin{equation}\label{eq:convergence5}
\begin{split}
    \Big| \mathbb{E} \big[ {\|{\nabla} f({\bf W}_{L})\|} \big] & - \mathbb{E} \big[ {\|{\nabla} f({\bf W}^*)  \|} \big]\Big|  \leq (1-\delta)^L(1-{\epsilon})^L \ \mathbb{E} {\|  {\nabla} f({\bf W}_{0})\|} + \frac{1}{\epsilon} \left( \zeta(Q, \delta) + \frac{\delta M}{1-\delta} \right),
\end{split}
\end{equation}
which completes the proof.
\end{proof}

\newpage
\section{Additional Experiments}\label{app:ExtraExp}
In this section, we complement our discussions with an ablation study of the descending constraints.

\subsection{Ablation study} 
\begin{figure}[h]
    \centering
    \includegraphics[width=0.7\textwidth]{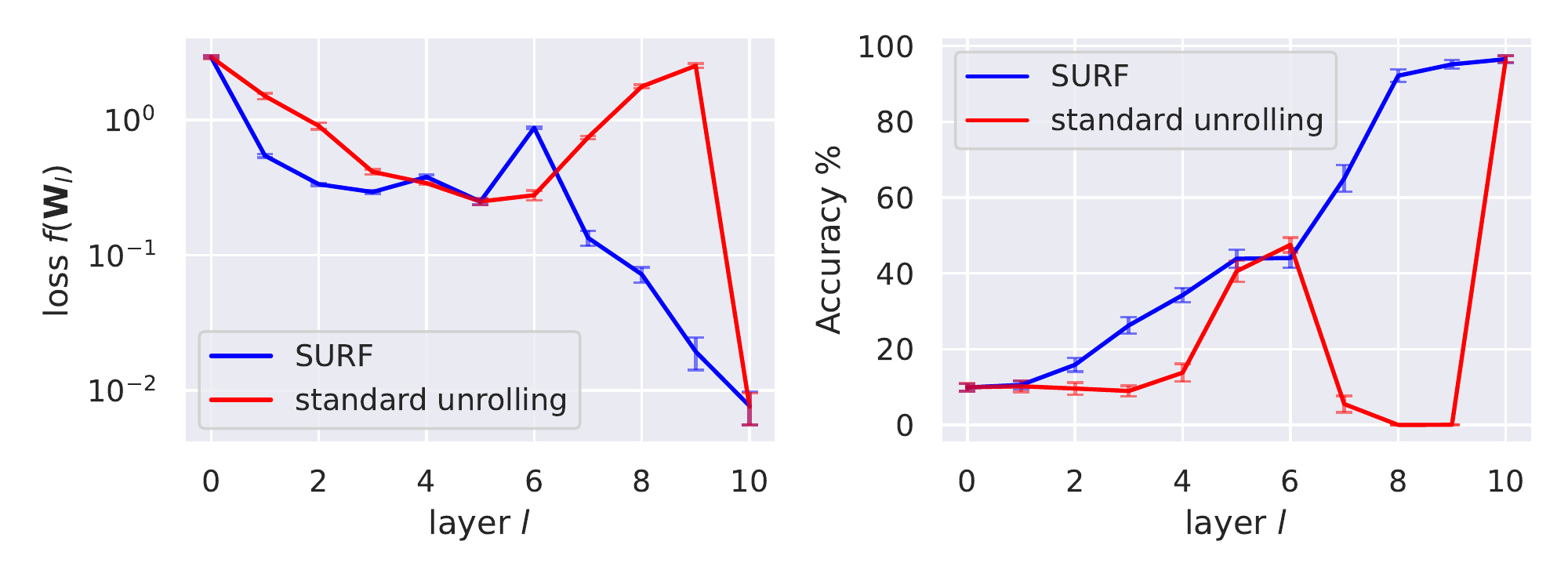}
    \caption{\textbf{Convergence Guarantees.} Comparison of the loss and accuracy (evaluated over 30 test datasets sampled from MNIST) with and without the constraints in \eqref{eq:surf} across the unrolled layers of U-DGD. Observe that SURF converges gradually to the optimal.}
    \label{fig:Ablation}
\end{figure}

To assess the effects of the descending constraints on the training performance, we compare the test loss and accuracy with and without these constraints in Figure \ref{fig:Ablation}.
We trained both models over meta-training dataset construced from MNIST dataset \cite{MNIST}. The set-up of the experiment is identical to the one of the original experiment in \cref{sec:results}. The network among the agents is chosen here as a $3$-degree regular graph.
The figure shows that the unrolled optimizer trained using SURF, depicted in blue, converges gradually to the optimal loss/accuracy over the layers. However, the standard unrolled optimizer trained without the descending constraints failed to maintain a similar behavior even though it achieves the same performance at the final layer. In fact, the accuracy jumps from $0\%$ to $96\%$ at the last layer,
which would make the optimizer more vulnerable to additive noise and perturbations in the layers' inputs, as we show in the following experiment.

\begin{figure}[b]
    \centering
    \includegraphics[width=0.4\linewidth]{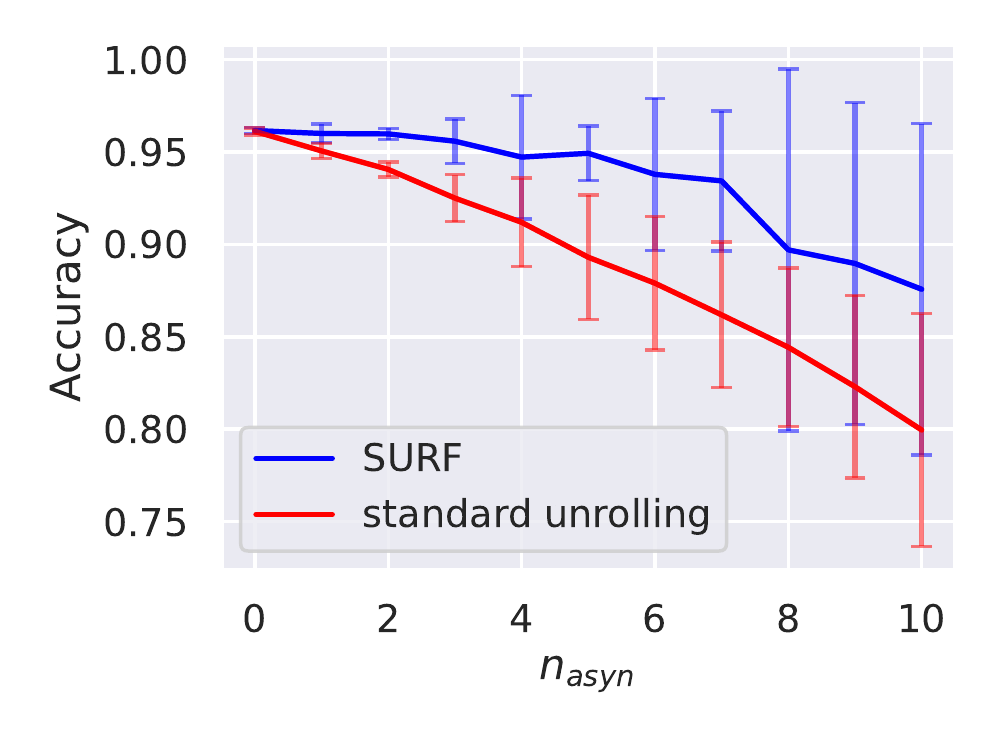}
    \caption{\textbf{Asynchronous Communications.} Comparison of the test loss and accuracy in different communication environments where $n_{asyn}$ agents are asynchronous with the rest of the network.}
    \label{fig:Asyn}
\end{figure}

The significance of having convergence over the layers to the optimal lies in the optimizer's response to perturbations. Standard algorithms, such as DGD, persist in moving toward the minimum even after their trajectories are perturbed by noise. If unrolled optimizers lack this feature, their resilience against perturbation is jeopardized, as reported by \cite{hadou2023robust}. Consequently, SURF, through its descending constraints, endows U-DGD with robustness to perturbations. To assess this quality, we consider one form of perturbation that occurs in asynchronous settings during inference. In this setting, $n_{asyn}$ randomly chosen agents fail to update and send their estimates simultaneously with the rest of the agents, and, therefore, outdated versions communicated at previous layers of their estimates are utilized by their neighbors. This would create a change in the input distribution at each layer, affecting the performance of the neural optimizers--a phenomena observed in machine learning models in general. The distribution shift is more influential when the change in the estimates from one layer to its successor is notable. This implies that a gradual change in the estimates across the layers, similar to the one achieved by SURF, would help mitigate the effect of asynchronous communications.

To assess the robustness of U-DGD to these purterbations, we evaluate the two U-DGDs trained above in this asynchronous setting and report their performance
in Figure \ref{fig:Asyn}. The figure shows that our constrained method SURF is more resilient, as the deterioration in the performance is notably slower than that of the case with no constraints.

\subsection{Hyperparameters}
\textbf{Decentralized FL benchmarks.}
In both DGD and DSGD, the agents update their estimates based on their local data through one gradient step at each communication round. The gradients in DGD are computed over a mini-batch of $10$ data points/agent compared to one data point in DSGD. In DFedAvgM, each agent takes $6$ gradient steps with momentum at each communication round. The step sizes are $10^3$, $10^4$, $10^2$ in DGD, DSGD, and DFedAvgM, respectively.

During evaluation, we use a meta-testing dataset that consists of $30$ downstream datasets. Similar to the process outlined in \cref{fig:unrolled_FL}, the training examples of the downstream dataset are used to train the softmax layer, and subsequently, the test accuracy is  computed over the testing examples.
The test accuracy is then averaged over the $30$ datasets and is reported in \cref{fig:speed}.


\end{document}